\pdfoutput=1

\documentclass[11pt]{article}

\usepackage[]{ACL2023}

\usepackage{times}
\usepackage{latexsym}
\usepackage{amssymb}
\usepackage{booktabs}
\usepackage{tabularx}
\usepackage{makecell} 
\usepackage[cmyk]{xcolor}
\usepackage{longtable}
\usepackage[T1]{fontenc}
\usepackage{subfig}
\usepackage{graphicx}
\usepackage[utf8]{inputenc}
\usepackage{multirow}
\usepackage{microtype}

\usepackage{inconsolata}
\usepackage{CJKutf8}
\definecolor{myorange}{cmyk}{0,0.5,1,0}
\definecolor{mylime}{cmyk}{0.65, 0, 0.85, 0.05} 

\definecolor{myyellow}{cmyk}{0, 0.15, 1, 0} 

\definecolor{mypink}{cmyk}{0, 0.68, 0.25, 0} 

\definecolor{mybrown}{cmyk}{0, 0.15, 1, 0.5} 
\title{Creative Convergence or Imitation?  Genre-Specific Homogeneity in LLM-Generated Chinese Literature}

\author{Yuanchi Ma, Kaize Shi, Hui He, Zhihua Zhang, Zhongxiang Lei, Ziliang Qiu, Renfen Hu, Jiamou Liu }

\begin{document}
\maketitle
\begin{abstract}

Large Language Models (LLMs) have demonstrated remarkable capabilities in narrative generation. However, they often produce structurally homogenized stories, frequently following repetitive arrangements and combinations of plot events along with stereotypical resolutions.
In this paper, we propose a novel theoretical framework for analysis by incorporating Proppian narratology and narrative functions. This framework is used to analyze the composition of narrative texts generated by LLMs to uncover their underlying narrative logic. Taking Chinese web literature as our research focus, we extend Propp's narrative theory, defining 34 narrative functions suited to modern web narrative structures. We further construct a human-annotated corpus to support the analysis of narrative structures within LLM-generated text.
Experiments reveal that the primary reasons for the singular narrative logic and severe homogenization in generated texts are that current LLMs are unable to correctly comprehend the meanings of narrative functions and instead adhere to rigid narrative generation paradigms.
\end{abstract}

\section{Introduction}
The intersection of AI and creative writing has witnessed rapid evolution, with Large Language Models (LLMs) now demonstrating the ability to produce fluent, grammatically coherent stories at scale \cite{gpt,guo2025pen}. This generative capability is particularly disruptive for high-volume literary markets, such as Chinese Online Literature (web fiction) \cite{marco2024pron}. As of 2024, this industry generated approximately USD 6.9 billion in revenue and served nearly a billion readers globally\footnote{\scriptsize https://www.reuters.com/markets/emerging/chinas-online-literature-expands-overseas-readership-cultural-export-push-2025-06-17/}. Beyond its market scale, this genre offers a unique technical vantage point: it is characterized by highly serialized formats and distinct, trope-driven architectures such as the ``cultivation arcs'' in Xianxia or specific role dynamics in Urban romance \cite{jin2022analysis,100}. These structural constraints make the genre an ideal, rigorous testbed for evaluating whether AI models can master complex narrative patterns beyond mere sentence-level fluency.

Despite these advancements and the evident structural richness inherent in genres like Chinese web fiction, a critical limitation persists: LLM-generated narratives are frequently criticized for being ``homogenous'' or ``formulaic'' \cite{marco2024pron}. While the text can be fluent and grammatically correct on a sentence-by-sentence basis, the overarching stories often feel repetitious and lack genuine engagement over extended contexts. Existing diagnostic approaches have primarily focused on surface-level linguistic features, such as lexical diversity or $n$-gram repetition \cite{xu2025echoes}. However, we contend that these metrics fail to capture the fundamental issue. A compelling story transcends its individual words; it is intrinsically a dynamic sequence of causally linked events and character interactions. The perceived `boredom' and lack of novelty in AI-generated fiction thus derive not from word choice, but from a deeper ``structural collapse'', which manifests as a rigid adherence to simplistic plot templates that hinder meaningful narrative development and lead to predictable, uninspired outcomes.

Diagnosing this structural collapse proves challenging given the limitations of current quantitative frameworks in NLP for evaluating `plot logic' or narrative coherence. Standard metrics like BLEU or perplexity, while effective for surface-level linguistic evaluation, are ill-equipped to assess whether a protagonist's journey makes internal narrative sense or if a plot twist is logically earned within the broader story arc. To bridge this critical gap, we turn to the well-established domain of Structuralist Narratology, drawing specifically from the seminal work of Vladimir Propp \citet{po1}. While Propp's original morphology was developed through the analysis of Russian folklore, its fundamental premise that diverse stories can be reduced to a sequence of invariant 'narrative functions' offers a powerful and domain-agnostic lens for the computational analysis of plot structure \cite{po_allareas3,po2}.

In this paper, we introduce a general framework to diagnose structural homogenization in AI narratives by rigorously operationalizing narratology for the modern age. Our comprehensive approach encompasses three major innovations. First, moving beyond the traditional scope of folklore, we introduce a refined narrative-function taxonomy comprising 34 specific functions carefully adapted from Propp's original set to reflect the unique structural characteristics of contemporary Chinese online fiction. This adaptation is critical given the distinctive narrative conventions and structural properties of this genre, which have been explored in prior literary analyses \cite{jin2022analysis}. Our taxonomy thus incorporates genre-specific elements such as The Golden Finger, Face-Slapping, and Transfiguration that are prevalent in modern web novels but absent from traditional folktales. Second, to provide robust empirical grounding for our analysis, we construct a large-scale, function-aligned corpus containing 1.0k expertly annotated narrative segments derived from 100 representative novels across five major genres \cite{100}. This manually curated dataset, following the precedent of previous efforts to build narrative function corpora in NLP \cite{po_tag2}, serves as a critical ground truth for both analyzing existing narrative structures and evaluating AI models' understanding of them.
Our main contributions are as follows:
\smallskip
\begin{itemize}
    \item A redesigned set of 34 narrative functions adapted from Propp, tailored to Chinese web fiction.
    \smallskip
    \item A function-aligned corpus of Chinese web novels annotated by experts. It contains 1.0k pieces of labeled data.
    \smallskip
    \item An evaluation framework for analyzing the content generated by LLMs can directly analyze the internal logic of the generated content.
\end{itemize}
\section{Related Works}

\subsection{ Propp's Narratology Theory in NLP.} Early efforts to annotate narrative function symbols on narrative data were initiated by~\citet{po_tag1}, and later extended by~\citet{po_tag2}.  They manually annotated 15 Russian folk tales from the original Propp corpus, amounting to a total of 18,862 words.  
\citet{po_allareas4}  emphasized multi-level annotation, i.e., grammar, semantics, and narrative structure, while laying the groundwork for developing machine learning algorithms capable of extracting culturally embedded plot structures from folk tale corpora.
More recent study~\cite{po3,po2} leveraged LLMs to automate the annotation of narrative function symbols directly from story outlines.  However, these approaches primarily focus on capturing the full range of narrative elements defined by Propp for the purpose of structural analysis, rather than enabling text generation tasks.  As such, they fall short in addressing the problem of narrative diversity and content homogenization in LLMs.
\subsection{Creative writing}

Creative writing is a sophisticated and intrinsically human endeavor that highlights self-expression and introspection. It is widely recognized as an art form that prioritizes emotional expression~\cite{perfer1,method2}, originality, and creativity. In particular, various themes in the literary domain serve as a medium for exploring and revealing unique human insights. Existing studies further indicate that LLMs continue to trail behind top human authors in creative writing competitions~\cite{method1}. However, in the realm of Chinese online literature, the demand for high-quality writing is less stringent, allowing LLM-generated writing to occupy a broader space.

\begin{figure*}[t]
	\centering
	\includegraphics[width=1\linewidth]{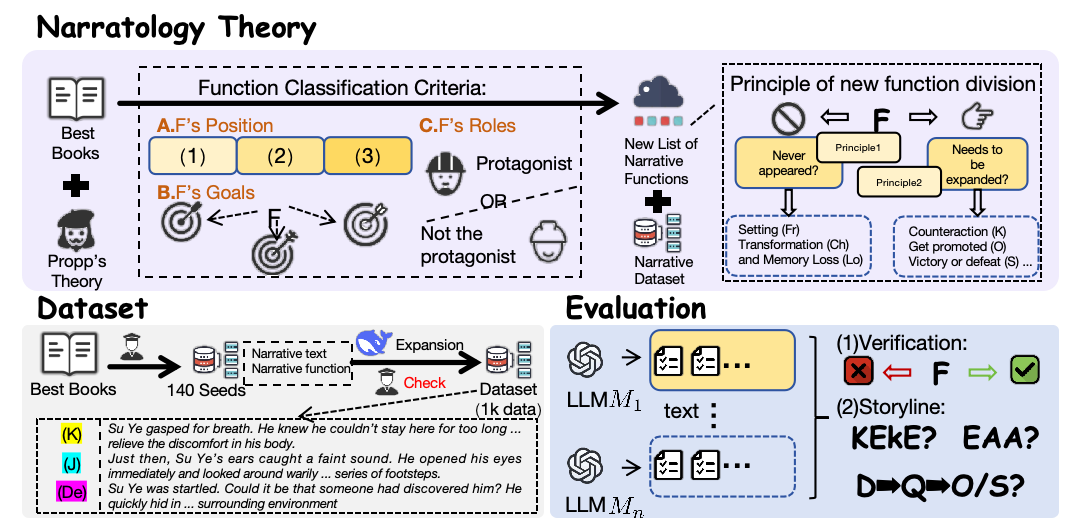}
	\caption{The working framework consists of theoretical, dataset and evaluation. The theoretical section presents our two principles for dividing narrative texts, including the fundamental division principle and the new function improvement principle.}
	\label{workflow}
\end{figure*}

\section{Proppian Taxonomy for Chinese Web Fiction}
Our analytical framework consists of two components. First, drawing on Propp’s narratological classification of narrative functions, we analyzed the most influential works in Chinese online literature using the narrative function segmentation method proposed in Propp’s original theory, and derived a narrative function taxonomy suitable for modern online literature. Second, based on the revised classification criteria, we constructed a large-scale dataset to support narrative function analysis of texts generated by LLMs. The overall workflow is illustrated in Figure~\ref{workflow}.

\subsection{Theoretical Induction Method}
\begin{figure}[t]
\centering
	\includegraphics[width=1\linewidth]{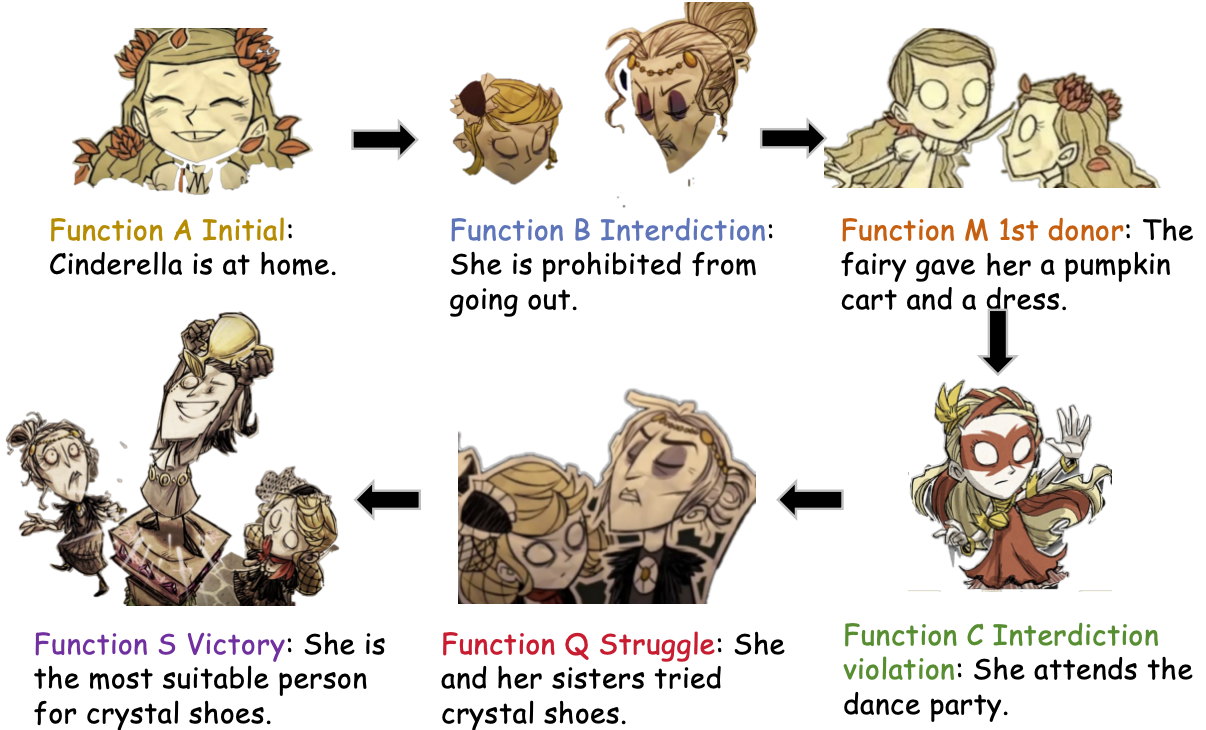}
\caption{An overview of the Cinderella. This picture analyzes the entire storyline and the related narrative functions.}
\label{ex2}
\end{figure}

Propp's structuralist narratology offers an elegant foundation for narrative scaffolding: based on the analysis of approximately 100 Russian folktales, he proposed 31 invariant {\em narrative functions}, i.e., plot-building units such as interdiction, villainy, departure, and return, that recur in fixed order across diverse stories, abstracting away from character or setting \cite{po1,po2} (see Appendix A.2). For instance, the classic fairy tale Cinderella can be simply abstracted as a permutation and combination of six narrative functions in Figure~\ref{ex2}. Although his taxonomy has deeply influenced literary theory and storytelling frameworks, it was conceived within early 20th‑century cultural contexts and does not fully align with modern, serialized genres like Chinese online fiction, which is dominated by repeated tropes and structural templates rather than folklore-style arcs.

\begin{table*}[t]
  \centering
  \resizebox{\linewidth}{!}{
  \small  
  \setlength{\tabcolsep}{0.5mm}
  \begin{tabular}{llll}
  \toprule
   Function (symbol) &Narrative function description & Function (symbol) &Narrative function description\\
   \midrule
    Initial situation (A) & The initial scene & Interdiction (B) & A prohibition is imposed on the hero\\
    Interdiction violation (C) & The interdiction is violated & Reconnaissance (D) & The enemy  reconnaissance\\
    \textbf{Delivery (E)} & \textbf{The character obtains intelligence/response} & \textbf{Trickery (F)} & \textbf{Foreshadowing/ traps/suspense/hints}\\
    Complicity (G) & The victim submits to deception& \textbf{Villainy} (H) & \textbf{The enemy's crime/ambition}\\
    Lack (I)& The scarcity of the protagonist & Mediation (J)& Misfortune is made known \\
    \textbf{Counteraction (K)}&\textbf{The protagonist's response/psychological} & \textbf{Departure (L)} & \textbf{Protagonist on the journey}\\
    \textbf{1st donor (M)} & \textbf{The giver or the golden finger appears }& \textbf{Hero’s reaction (N)} & \textbf{The hero reacts to the giver} \\
    \textbf{Get promoted (O)} &\textbf{Acquires items or enhances his abilities} &Transfer (P) &Spatial transfer \\
    Struggle (Q)& The hero and enemy engage in direct conflict & Marking (R) & The hero is marked or identified \\
    \textbf{Victory or defeat (S)}& \textbf{The victory or defeat of the character} &Liquidation of lack (T) & Villainy or lack resolved \\
    Return and Pursuit (U)& The hero returns. The enemy chases the hero & \textbf{Transformation (Ch)} & \textbf{Change in role or power relationship}  \\
    Rescue (V) & The hero is saved from pursuit or danger &Unrecognized arrival (W) & Protagonist is not recognized \\
    \textbf{Setting (Fr)}& \textbf{the strength/Ability system setting}  & &either actively or passively\\
    Unfounded claims (X)& The protagonist is confronted with unreasonable & \textbf{Transfiguration (Fa)} & \textbf{The protagonist's beautification,}\\
    & demands, unfair competitions or difficulties & &\textbf{casual clothes or pseudonyms}\\
    Solution (Z)& The task is accomplished or resolved & Recognition (Re)& The hero is recognized/acknowledged \\
    Exposure (De) & The enemy’s identity or deception is exposed & Difficult task (Y) & The hero is given a difficult task\\
    \textbf{Emotion (Em)} & \textbf{The changes of the characters' emotions}& \textbf{Beyond (Fi)}& To complete or beyond unreasonable \\
    \textbf{Memory Loss (Lo)} & \textbf{The protagonist's active/passive memory loss } & &demands\\
    
  \bottomrule
  \end{tabular}
  }
  \caption{A list of narrative functions based on Chinese online novels, where the bold ones have been improved}
  \label{listf}
\end{table*}

To bridge this gap, we systematically analyzed 100 popular Chinese web novels spanning Fantasy, Xianxia, Romance, Time Travel, and Urban genres, curated via the National Library of China~\cite{100} (see Appendix A.5). Each novel contains 100,000 to 10 million+ words. These genres collectively reflect the core narrative mechanisms of contemporary Chinese digital storytelling. Drawing upon the classification framework proposed by \citet{fenlei2} and the principles of Propp, we have reorganized the narrative function for Chinese online literature. 
As a result, we identified a list $F$ of 34 narrative functions $f_i$, $i\in \{1,\ldots,34\}$, as shown in Table~\ref{listf}. 




As illustrated by the narrative function division criteria in Figure~\ref{workflow}, we determine the narrative function of a paragraph or sentence sequentially from three complementary aspects.
\textit{1) The \textbf{position} of sentences in an article}: In narrative texts, certain functions, such as the initial scene are more likely to occur at the beginning of the text. Moreover, the occurrence order of some functions is constrained. For example, an Interdiction (B) must precede an Interdiction violation (C).
\textit{2) The specific \textbf{goals} achieved by the function}: We examine whether the function advances the plot toward a particular narrative stage or objective. For instance, when the protagonist moves from one location to another, the narrative function fulfills the objective of spatial transfer and is categorized as Transfer (P).
\textit{3) Characteristics or behaviors of the involved \textbf{roles}}: Narrative functions are associated with the actions or states of specific role types, such as Counteraction (K) and Delivery (E). In our classification, unlike Propp’s original division of characters into seven role types, we focus on distinguishing between main and non-main characters, thereby avoiding potential character ambiguity caused by the large number of character archetypes in Chinese online literature.

By refining the narrative function division criteria derived from Propp’s original theory, while preserving the original narrative functions as much as possible, we provide updated interpretations and incorporate additional functions. In total, 15 narrative functions have been revised or newly introduced. These revisions target functions that are rare or absent in classical Russian folk tales but are prevalent in contemporary online literature.

First, certain narrative phenomena cannot be adequately categorized under Propp’s original framework, as they do not correspond to any existing function. Consequently, we define these phenomena as independent narrative functions. These newly defined functions include Setting (Fr), Transformation (Ch), and Memory loss (Lo). The remaining 12 functions extend and reinterpret Propp’s original functions through supplementary explanations.
For example, in Propp’s original theory, the First donor (M) denotes the appearance of a benefactor. In Cinderella, the fairy godmother provides magical assistance, such as the pumpkin carriage and the crystal shoes. In Chinese online literature, however, this role is commonly described as a “golden finger,” referring to mechanisms by which characters are endowed with extraordinary abilities, items, or advantages. Although the underlying narrative function remains consistent, an extended interpretation is required to align with contemporary literary conventions.

\begin{figure}[t]
\centering
	\includegraphics[width=1\linewidth]{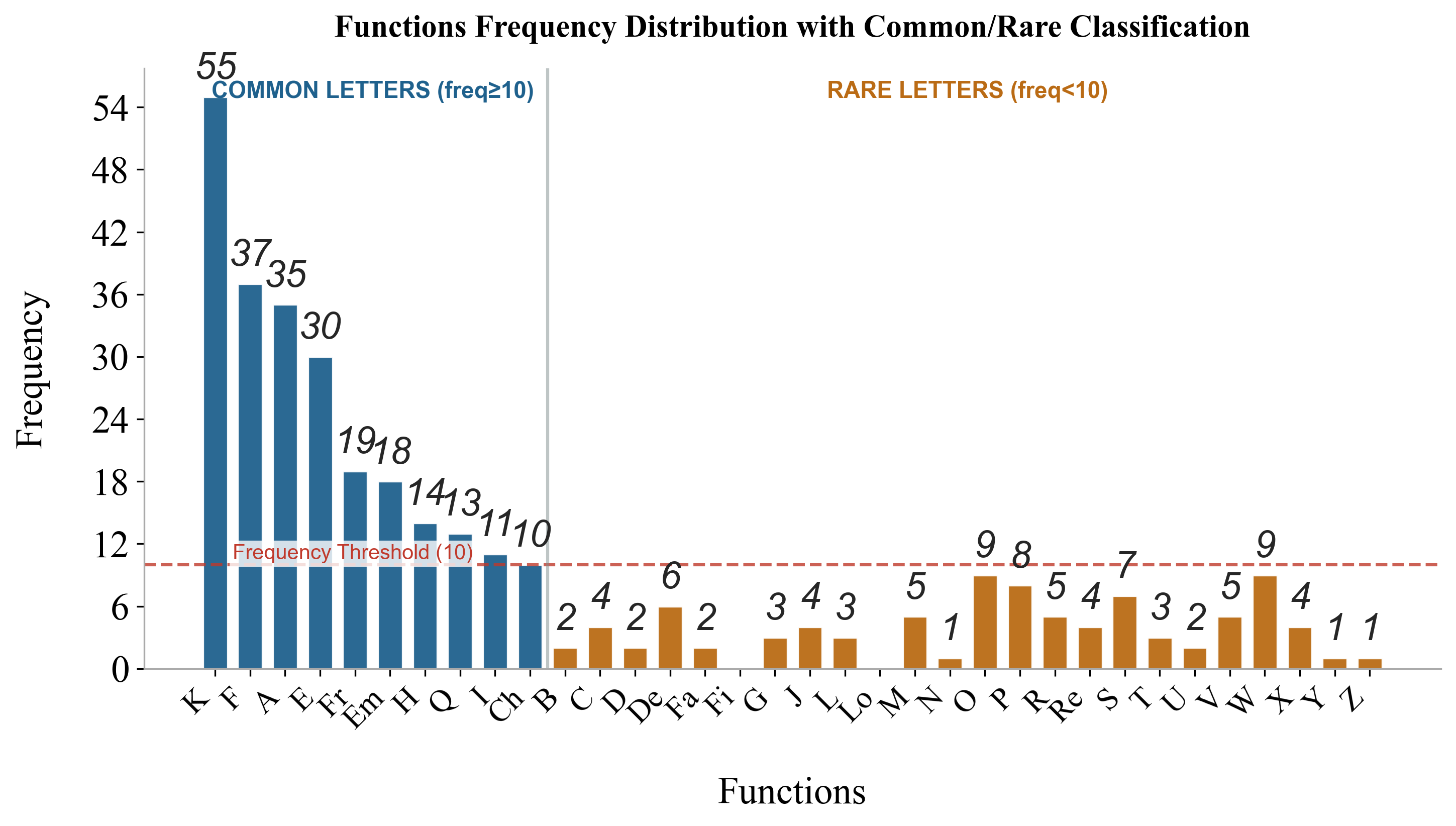}
\caption{Narrative function frequency sampling}
\label{listTJ}
\end{figure}

For the summarized narrative functions, we further analyze their frequencies of occurrence to identify which functions can be regarded as common types. We randomly divided 100 online novels into five groups, and within each group, four novels were randomly selected for sampling. For each selected novel, 2,000 characters were randomly sampled for narrative function analysis. According to our statistics, all narrative functions appeared a total of 332 times, with an average frequency of approximately ten occurrences per function. Narrative functions whose frequencies exceed the average are defined as common narrative functions, whereas the remaining functions are classified as uncommon. Figure~\ref{listTJ} presents detailed statistical results and indicates whether each narrative function is classified as common.

\subsection{Annotated Corpus Construction}



To support the analysis of the narrative function of the text, we have constructed a corpus containing narrative function annotations based on 100 Chinese online novels. Each sample includes: the original text segment and the function symbols annotated by experts. 

Two expert annotators, including one professional web-fiction author and one genre critic, manually assigned function symbols to randomly selected text segments (paragraph-length units, free of thematic bias). For each narrative segment, they also write reasoning for the task output. Inter‑annotator agreement $\kappa \approx 0.83$, indicating high reliability in symbolic labeling. 
The initial dataset contains 140 seeds, divided equally between the two tasks. Corpus statistics reveal balanced coverage of the 34 defined narrative functions (each appearing in at least 4 sample segments). The annotation dataset reflects a realistic distribution of narrative functions across genres and story structures. Among them, an example of dataset sampling is shown in Figure.~\ref{workflow} \{\textit{dataset}\}.

Starting from the initial seeds, we expanded the dataset once to obtain the entire dataset. In the expansion, we used Deepseek-R1 to expand the narrative function classification dataset, bringing the entire dataset to 1.0 k entries and enriching the narrative function samples of each category. Finally, through the review and adjustment by experts, it is ensured that all data samples are labeled correctly. The dataset is available at \textit{https://anonymous.4open.science/r/acl26-ED4E/}.

\section{Experimental}



In the experimental, we conducted a detailed analysis of the text generation capabilities and narrative function understanding of existing mainstream LLMs. The analysis is conducted from two complementary aspects:

{\bf(1) Verification of comprehension ability:}  Select fixed online novel texts from our labeled dataset and use each baseline model to analyze their narrative function composition, thereby evaluating the understanding ability of each baseline for the narrative texts. In addition, we also separately calculated the recognition rate scores of LLMS for common or uncommon narrative functions. This analysis determines whether LLMs can correctly understand and distinguish different narrative functions.

{\bf(2) The discovery of the storyline in narrative texts:}
Discovering storyline structures in narrative texts is essential for understanding narrative generation mechanisms. Since the training data of LLMs mainly comes from human-created texts, an in-depth analysis of the internal composition of narrative functions (i.e., storylines) in human texts can help reveal the internal logic of the texts generated by LLMs and identify the common paradigms of the same plots in different types of narrative texts. We analyze diverse types of narrative texts using both LLM-based analysis and manual review to explore the underlying chains of narrative logic embedded within these texts. Identifying these internal patterns not only facilitates a deeper understanding of the homogenization problem in LLM-generated narrative texts, but also provides valuable insights for humanities and social science research.
\smallskip
\subsection{Experimental Setup}
This section mainly introduces the baseline and evaluation. For more evidence on the configuration of expert teams and the phenomenon of homogenization, please refer to Appendix A.1.

\noindent \textbf{Baselines.}
To ensure the robustness and comprehensiveness of our evaluation, we selected GPT~\cite{gpt} and six state-of-the-art Chinese large language models (LLMs) as baselines. These include Doubao, Qwen3~\cite{qianwen}, Xuanyuan~\cite{xuanyuan}, Qianfan, Kimi, and Deepseek~\cite{deepseek}.

\smallskip

\noindent \textbf{Evaluation Metrics.}
For the assessment of the accuracy of LLMs in understanding narrative functions, we make judgments by calculating the\textbf{ ACC, Recall}, and \textbf{F1} values of their predictive narrative functions. In addition, for the narrative composition of the text, when analysis is needed, we adopt the annotation method, marking the corresponding narrative function symbols after the corresponding text.

\subsection{Experimental Result}

\noindent \textbf{Verification of comprehension ability.}
In this experiment, we selected narrative paragraphs written by two human authors, which collectively contain 10 narrative functions drawn from the constructed dataset. Given a baseline set of 34 narrative functions, each baseline model was tasked with identifying the narrative functions present in the text. This procedure is repeated over 10 independent rounds, with five predictions generated in each round. Finally, the mean performance and standard deviation of each baseline are computed. The selected texts include four common narrative functions ((K), (E), (F), (A)) and six rare narrative functions ((De), (Fa), (Lo), (J), (Re), (G)). All text fragments used in this experiment are presented in Table~\ref{Otext}.

According to the experimental results reported in Table~\ref{listf2}, Qwen3 and Doubao achieve the highest performance in recognizing and understanding narrative functions, with an accuracy of 0.364. Notably, Qwen3 attains a recognition accuracy of up to 60\% on common narrative functions, suggesting a stronger capability to capture frequently occurring narrative segments with relatively explicit structural patterns. These functions typically exhibit more salient surface-level linguistic cues. By contrast, Doubao achieves a recognition accuracy of 33.3\% on rare narrative functions, which may indicate stronger potential in generalization, sensitivity to low-information signals, or the handling of more abstract and implicit narrative elements. Overall, most LLMs are able to recognize only a limited subset of common narrative functions (e.g., character response (E)), while exhibiting substantially lower performance on uncommon functions. This result indicates that none of the evaluated baselines can accurately comprehend our newly proposed narrative function framework, even when the original Proppian definitions are retained. This further suggests that existing LLMs lack a clear conceptual understanding of narrative theory and, more fundamentally, fail to grasp the structured, role-driven, and intent-oriented abstraction underlying the notion of a narrative function.

\begin{table}[h!]
	\centering
    \small
	\begin{tabularx}{0.5\textwidth}{X}
		\toprule
		   Narrative text generation\\
        \midrule
        \begin{CJK}{UTF8}{gbsn}
        苏业倒吸了一口气。他知道他不能在这里呆太久。他需要尽快恢复体力，继续寻找出路(K)。就在这时，苏叶的耳朵里听到了一个微弱的声音(J)。苏叶吓了一跳。会不会是有人发现了他？他赶紧躲在石台的阴影里，屏住呼吸，等待着有人的到来(K)。最后，一个人影出现在裂缝中。那是一个人，披着一件黑斗篷，脸上戴着面具，使他的脸模糊不清。他手里拿着火炬，照亮了周围的环境（De）。黑衣人似乎没有注意到苏叶的存在。他环顾四周，好像在寻找什么东西。他动作敏捷，显然不是一个普通人（E)。突然，黑衣人脱下黑外套，露出英俊的脸庞(Fa)。看到他的那一刻，一股巨大的能量波动涌进苏的大脑，使他瞬间失去了记忆(Lo)。
        \end{CJK}\\
        \midrule
        Su Ye gasped for breath. He knew he couldn't stay here for too long. He needs to regain his strength as soon as possible and continue to look for a way out.\colorbox{pink}{(K)}
        Just then, Su Ye's ears caught a faint sound. \colorbox{yellow}{(J)}. Su Ye was startled. Could it be that someone had discovered him? He quickly hid in the shadow of the stone platform, held his breath and waited for the arrival of someone. \colorbox{pink}{(K)} Finally, a figure emerged in the crack. It was a person wearing a black cloak and a mask on his face, making his face indistinct. He was holding a torch in his hand, illuminating the surrounding environment. \colorbox{yellow}{(De)}
        The man in black seemed not to have noticed Su Ye's presence. He looked around as if searching for something. He was very agile in his movements and clearly was no ordinary person. \colorbox{pink}{(E)}
        Suddenly, the man in black took off his black coat, revealing his handsome face. \colorbox{yellow}{(Fa)}
        The moment he saw him, a huge energy fluctuation rushed into Su’s brain, causing him to lose his memory instantly. \colorbox{yellow}{(Lo)}\\
        \midrule
        \begin{CJK}{UTF8}{gbsn}四面巍峨的高台上，人声鼎沸（A）.王释影在台上微笑说道：“早就听说夕慈镇出了一个少年英雄，今日一见果然不同凡响(Re)  。王释影装作不认得他，月城武也不说破(G) .王释影微笑着鼓励了几句,便宣布明日一早，六派的仙岛将会飞临心辰殿上空(F)。\end{CJK}\\
        \midrule
         On the lofty platforms on all sides, the sound of people was deafening\colorbox{pink}{(A)}. Wang Shiying smiled on the stage and said, "I've long heard that there's a young hero from Xici Town. Today, seeing him, he's truly extraordinary.\colorbox{yellow}{(Re)}"  Wang Shiying pretended not to recognize him, and Yuecheng Wu didn't say anything either \colorbox{yellow}{(G)}. Wang Shiying smiled and encouraged him a few words, then announced that the immortal islands of the six sects would fly over the Xinchen Hall early the next morning \colorbox{pink}{(F)}.\\
        \bottomrule
	\end{tabularx}
	\caption{The narrative text being analyzed}
    \label{Otext}
\end{table}%

\begin{table*}[t]
  \centering
  \small
  \resizebox{\linewidth}{!}{
  \begin{tabular}{llll|ll}
  \toprule
   Model & ACC $\uparrow$ (Common/Rare) & Recall $\uparrow$ (Common/Rare) & F1 $\uparrow$ (Common/Rare) & ACC $\uparrow$ (Sum) & F1 $\uparrow$ (Sum)\\
   \midrule
   GPT-4o  & 0.200($\pm 0.0$)/0.167($\pm 0.2$) & 0.200($\pm 0.0$)/0.167($\pm 0.2$) & 0.167($\pm 0.0$)/0.167($\pm 0.2$) & 0.182($\pm 0.2$) & 0.167($\pm 0.2$)\\
   GPT-4oMini & 0.200($\pm 0.0$)/0.000($\pm 0.0$) & 0.200($\pm 0.0$)/0.000($\pm 0.0$) & 0.167($\pm 0.0$)/0.000($\pm 0.0$) & 0.091($\pm 0.0$) & 0.167($\pm 0.0$)\\
    Qwen3-8b & 0.200($\pm 0.0$)/0.000($\pm 0.0$) & 0.200($\pm 0.0$)/0.000($\pm 0.0$) & 0.167($\pm 0.0$)/0.000($\pm 0.0$) & 0.091($\pm 0.0$) & 0.167($\pm 0.0$)\\
    Qwen3-32b & \textbf{0.600($\pm 0.0$)}/0.167($\pm 0.2$) & \textbf{0.625}($\pm 0.0$)/0.167($\pm 0.2$) & \textbf{0.667}($\pm 0.0$)/0.167($\pm 0.2$) & \textbf{0.364}($\pm 0.2$) & 0.282($\pm 0.2$)\\
     Qianfan & 0.400($\pm 0.2$)/0.167($\pm 0.0$) & 0.400($\pm 0.2$)/0.167($\pm 0.0$) & 0.400($\pm 0.2$)/0.167($\pm 0.0$) & 0.273($\pm 0.2$) & 0.276($\pm 0.0$)\\
      Doubao-pro & 0.400($\pm 0.0$)/\textbf{0.333($\pm 0.1$)} & 0.400($\pm 0.0$)/\textbf{0.333}($\pm 0.1$) & 0.570($\pm 0.0$)/\textbf{0.500}($\pm 0.1$) & 0.364($\pm 0.1$) &\textbf{0.533}($\pm 0.1$)\\
       Xuanyuan & 0.200($\pm 0.0$)/0.000($\pm 0.0$) &0.200($\pm 0.0$)/0.000($\pm 0.0$) & 0.200($\pm 0.0$)/0.000($\pm 0.0$) & 0.091($\pm 0.0$) & 0.200($\pm 0.0$)\\
        Kimi-v1 & 0.200($\pm 0.2$)/0.167($\pm 0.1$) & 0.250($\pm 0.0$)/0.167($\pm 0.0$) & 0.167($\pm 0.0$)/0.167($\pm 0.0$) & 0.182($\pm 0.2$) & 0.167($\pm 0.1$)\\
         Deepseek-V3 & 0.400($\pm 0.2$)/0.167($\pm 0.1$) & 0.500($\pm 0.0$) /0.167($\pm 0.0$) & 0.292($\pm 0.0$)/0.167($\pm 0.0$) & 0.273($\pm 0.2$) & 0.217($\pm 0.1$)\\
         Deepseek-R1 & 0.400($\pm 0.0$)/0.167($\pm 0.0$) & 0.400($\pm 0.0$)/0.167($\pm 0.0$) & 0.400($\pm 0.0$)/0.167($\pm 0.0$) & 0.273($\pm 0.0$) & 0.276($\pm 0.0$)\\

  \bottomrule
  \end{tabular}}
  \caption{Narrative function recognition results}
  \label{listf2}
\end{table*}

\noindent \textbf{The discovery of the storyline in narrative texts.}
To verify whether the same plot text has the same narrative development sequence, we selected six common plots in online literature for narrative function analysis. This inspiration originated from the initial Propp theory \cite{po1}, which posits that certain similar plot developments share the same narrative functional paradigm. Discovering these new narrative paradigms existing in Chinese online literature is conducive to deepening readers' or humanists' grasp of these works.

The selected plot types are highly representative, including battles, emotional exchanges, difficult tasks, adventures, and depictions of daily life. These plot types appear across nearly all five mainstream novel genres. From a corpus of 100 Chinese online novels, we first randomly selected 60\% of the works as samples. After expert screening, six plot instances were randomly extracted from each selected novel, with each plot limited to no more than 1,000 words. Narrative functions within these plots were first identified using Qwen3, which demonstrated strong performance, and were subsequently manually reviewed and corrected. For each randomly sampled plot, we analyzed its narrative function composition and subsequently quantified the position, sequence, and frequency of occurrence of each narrative function. Among them, the proportion of each paradigm's occurrence has reached at least 60\% or more (Appendix A.4).

After statistical analysis of the sampled texts, we found that all six plot types exhibit default narrative function arrangement patterns, as summarized in Table~\ref{caseframework}.  Here, $\to$ denotes linear narrative development, whereas $\hookrightarrow$ denotes nonlinear development. For instance, in battle scenes, most plots follow a linear progression: the initial situation (A) appears first, followed by the struggle (Q), and finally the plot concludes with victory or defeat (S). However, in Chinese online literature, this process is frequently accompanied by get promoted (O). These narrative processes typically unfold in a linear pattern and follow a chronological order.

\begin{table}[t]
  \centering
  \small  
  \setlength{\tabcolsep}{1mm}
  \resizebox{\linewidth}{!}{
  \begin{tabular}{l|l|l}
  \toprule
   Plots & Path paradigm & Explanation\\
   \midrule
    Battle & $(A)\to(Q)\to\{O/S\}$ & Probing to victory/defeat.\\

    Emotional& $(Em)\hookrightarrow(Ch)$ & Emotion-driven change.\\

    Difficult task& $(Y)\hookrightarrow(Z)$ & Task appears/completed.\\

    Adventure & $(P)\hookrightarrow\{M/O\}$ & Acquires main ability.\\

    Pretending & $(W)\to(De)\hookrightarrow(S)$ & Underestimating to crush.\\

    Daily life & $(A)\hookrightarrow(Ch)$ & Scene-driven change.\\
  \bottomrule
  \end{tabular}}
  \caption{Analysis of the narrative function arrangement patterns of six types of plots.}
  \label{caseframework}
\end{table}

There are also nonlinearly developed plots, such as emotional narratives, which often take emotional fluctuations (Em) as the marker of the plot’s initiation and relationship transformation (Ch) as the marker of its conclusion. However, this process is nonlinear in nature: the story does not unfold according to a natural temporal sequence, but instead relies on deliberate segmentation or narrative jumps. Although this development approach still revolves around a central narrative thread, the plot remains fragmented and ambiguous, requiring readers to actively reconstruct the underlying logic. Consequently, there is no unified paradigm for the intermediate progression of such plots; similarities are observed only in their initial and terminal narrative markers.

\subsection{Case Study}
To further demonstrate the plot paradigms summarized in this study, we used Table~\ref{Otext} as the narrative preface and required each LLM to perform continuation experiments, followed by an analysis of the intrinsic narrative functions in the generated texts. As shown in Table~\ref{case1}, we selected the three models that achieved relatively high performance in narrative function understanding. Although the sequence in which these models realize narrative functions occasionally varies, their underlying narrative logic remains largely consistent, with a strong tendency toward battle-oriented storytelling.

\begin{figure}[t]
	\centering
	\includegraphics[width=1\linewidth]{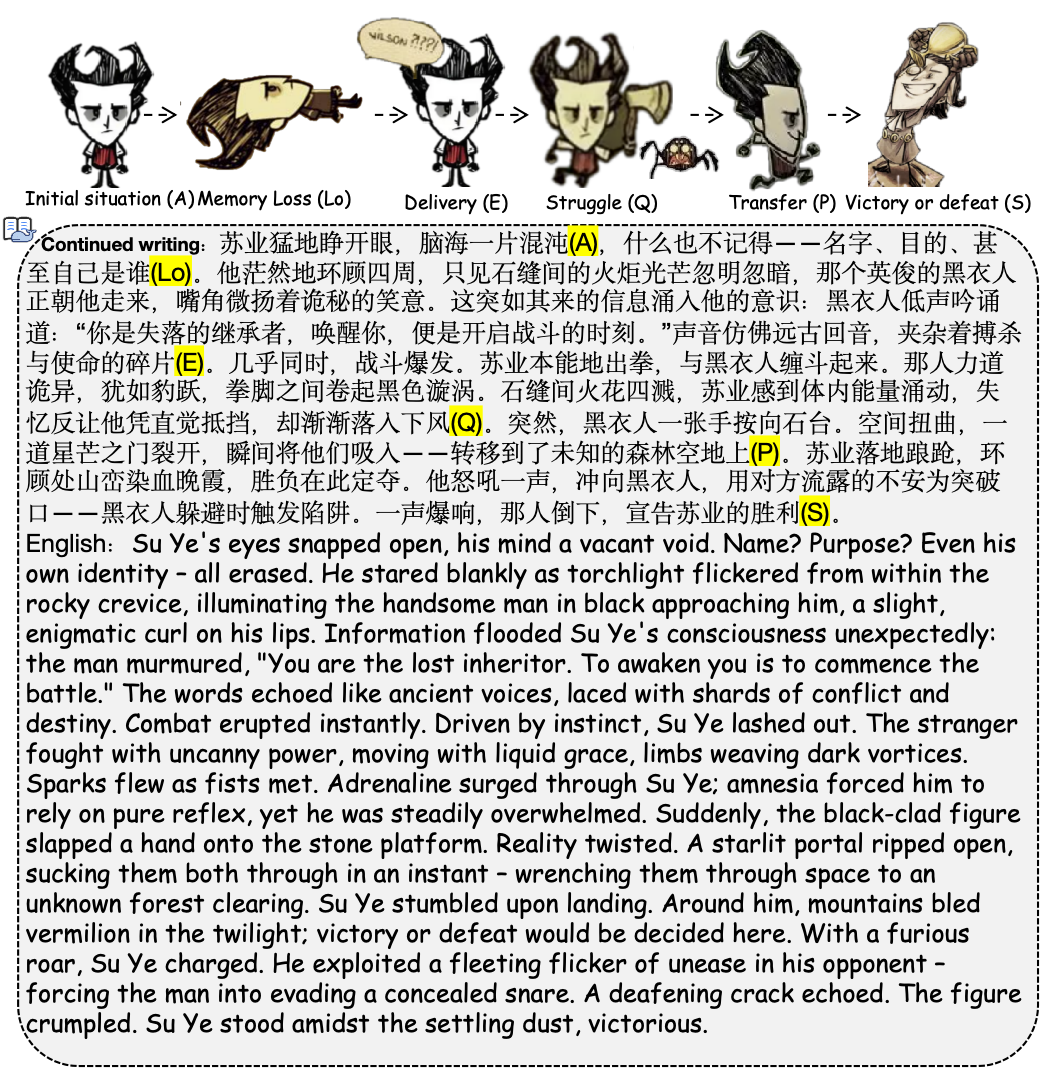}
	\caption{The narrative text generated by Deepseek in Ep.1 contains six narrative functions.}
	\label{casecon}
\end{figure}

Specifically, the narrative logic shared by the five continuation texts generated by Deepseek can be summarized as follows: initial situation (amnesia) (A/Lo) $\to$ information-driven development (E/J) $\to$ conflict escalation (H/Q) $\to$ acquisition of ability or marker (O/R) $\to$ battle outcome (S/P/U). This structure is consistent with the battle-plot narrative logic we have identified, in which (A), (Q), and (S) serve as key markers indicating the beginning, progression, and conclusion of the plot. As shown in Figure~\ref{casecon}, we visualize a representative story generated by Deepseek and provide the corresponding original outputs to illustrate this narrative pattern more intuitively.

\begin{table}[t]
  \centering
  \small  
  \setlength{\tabcolsep}{0.5mm}
  \resizebox{\linewidth}{!}{
  \begin{tabular}{llll}
  \toprule
   Ep. & Deepseek-R1 &Qwen3-32b & Doubao-pro\\
   \midrule
    1 & \colorbox{pink}{A}-Lo-E-\colorbox{pink}{Q}-P-\colorbox{yellow}{S} & \colorbox{pink}{A}-\colorbox{pink}{Q}-F-K-L-Fr-Lo  & \colorbox{pink}{A}-J-E-Lo-M-N-\colorbox{yellow}{O} \\
    2 & Lo-E-Em-P-\colorbox{yellow}{S}-U  & E-F-K-J-O-R-Re  & \colorbox{pink}{A}-J-E-Lo-M-N-\colorbox{yellow}{O}  \\
    3 & Lo-H-E-\colorbox{pink}{Q}-\colorbox{yellow}{S}-U  & A-J-E-\colorbox{pink}{Q}-P-\colorbox{yellow}{O}-R-Fr  & \colorbox{pink}{A}-J-E-\colorbox{pink}{Q}-M-N-\colorbox{yellow}{O}   \\
    4 & \colorbox{pink}{A}-F-\colorbox{pink}{Q}-O-R-F-\colorbox{yellow}{S}  & \colorbox{pink}{A}-F-K-J-\colorbox{yellow}{O}-R-W-Fr  & \colorbox{pink}{A}-J-E-\colorbox{pink}{Q}-M-N-\colorbox{yellow}{O} \\
    5 & \colorbox{pink}{A}-Lo-J-F-P-\colorbox{yellow}{S}  & E-F-K-\colorbox{pink}{Q}-L-R-\colorbox{yellow}{O}-Re  & \colorbox{pink}{A}-J-E-\colorbox{pink}{Q}-M-N-\colorbox{yellow}{O}  \\
  \bottomrule
  \end{tabular}}
  \caption{Analysis of the narrative function of the generated subsequent text based on the same preceding text.}
  \label{case1}
\end{table}

The texts generated by others (e.g.Doubao, see Appendix A.3) exhibit a high degree of content similarity, with the narrative functions of nearly all stories unfolding in an identical manner. This observation not only validates the correctness of the plot paradigms we have summarized, but also demonstrates that the fundamental cause of content homogenization lies in the arrangement and combination of narrative logic. Even when human-authored texts follow established plot paradigms, human writers tend to incorporate a wider range of narrative functions, thereby increasing textual diversity~\cite{perfer1}. As a result, human writing rarely exhibits the rigid arrangement and combination of narrative functions observed in LLM-generated texts. Furthermore, as shown in Figure~\ref{casecon}, the logical connections between narrative functions in LLM-generated texts are often weak, transitions are insufficiently smooth, and overall readability is reduced. This substantially degrades narrative quality and constitutes a primary factor underlying the gap between AI-generated and human-written texts.

Based on the aforementioned experiments, we conclude that the limited diversity in texts generated by existing LLMs stems fundamentally from their inability to comprehend narrative functions, which manifests as a rigid reproduction of conventional narrative templates. Specifically, we identified patterned combinatorial sequences of narrative functions as the structural origin of textual homogenization. This deficiency underscores two critical failures: (1) models cannot decode the semantic essence of individual narrative functions, and (2) They lack the ability to internalize the deep structural relationships among these functions, such as causality, logical progression or conflicts. Consequently, while LLMs proficiently replicate surface-level story elements (e.g., lexical choices or scene depictions), they remain blind to the axiomatic rules governing narrative skeletons, unable to dynamically model abstract story architectures. This compels models towards mechanically reproducing high-frequency "default templates" that align with stereotypical plot expectations, while systematically deactivating or attenuating functionally critical units (e.g., turning points like "violation of prohibitions") that demand structural reasoning. Such incapacity to internalize narrative mechanics constitutes the primary structural bottleneck: generated texts remain confined to predictable pathways devoid of strategic disruption, ultimately depriving narratives of organic complexity and inhibiting their evolutionary potential for rich, unexpected diversions.

\section{Conclusion}



In our paper we systematically analyzed the performance of existing LLMs in text generation by constructing a new narrative function schema and dataset based on Chinese web literature. Informed by our theoretical framework and dataset, we identified six prevalent narrative pattern types. While current LLM-generated texts adhere to these patterns, they fundamentally lack understanding of the inherent semantics of narrative functions. This profound deficit in structural cognition significantly impedes the ability of LLMs to generate diverse narrative texts with complex plot development.
Future research need focus on enhancing the model's understanding and reasoning capabilities regarding narrative structure, intention, causality, and functional semantics.

\section{Limitation}
This work is mainly focused on the field of Chinese online literature, with an emphasis on mainstream narrative genres. Although the proposed narrative function is effective in this case, our method may not be extended to other literary forms, such as poetry, experimental novels or genres with non-linear structures, because the narrative function is not very clear.

Furthermore, our research methods are limited to Chinese only. Applying this framework across languages may require reconstructing narrative classifications and datasets to explain language and cultural differences.

Although we analyzed 100 popular online novels, niche or avant-garde works have still not been fully represented. These texts may contain rare narrative structures that have not been captured in our current classification. Investigating such works can expand the scope of reporting and deepen the understanding of narrative diversity.

Finally, although we modeled the inter-function structure, the internal composition of each function remains unexplored. Future work may improve in-function modeling to support more detailed and expressive generation.

\bibliography{custom}
\bibliographystyle{acl_natbib}

\appendix
\section{Appendix A}
\subsection{Homogeneous region experiment and experts}
\label{sec:appendix1}

\begin{figure}[h]
\centering
\subfloat[Experimental setting]{
		\includegraphics[width=0.24\textwidth]{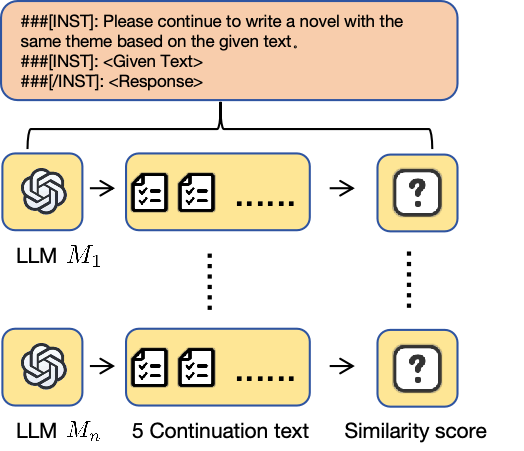}}
\subfloat[Experimental result]{
		\includegraphics[width=0.24\textwidth]{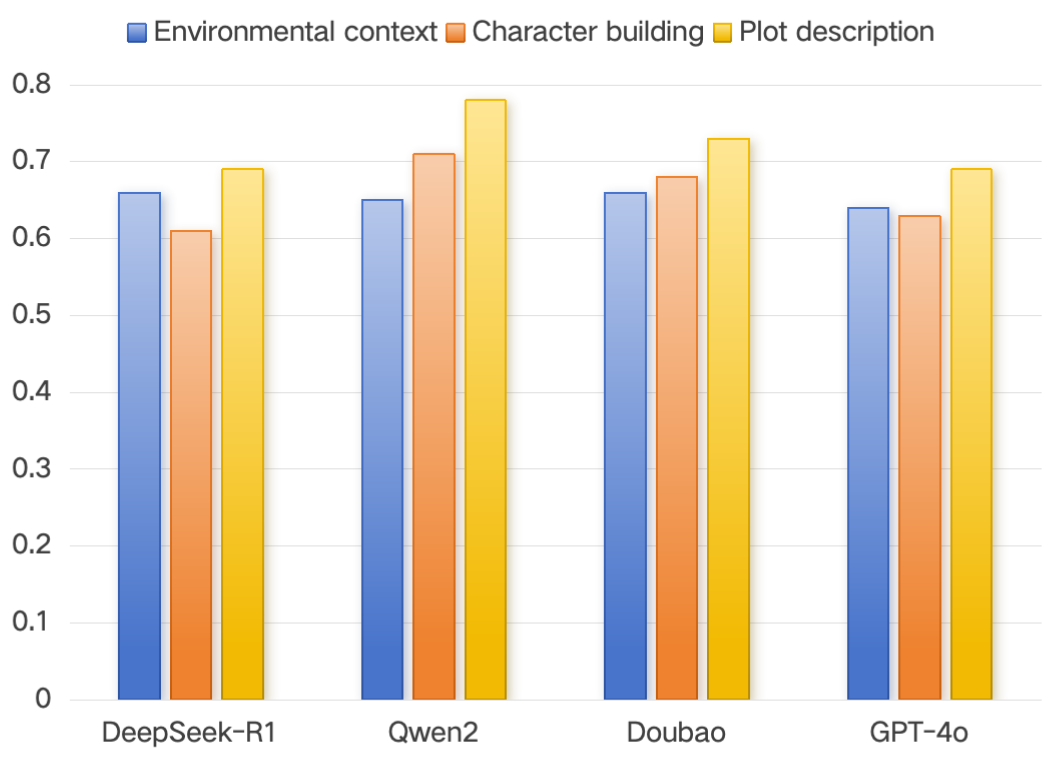}}
\caption{Our experimental approach is shown in (a). Based on the pretexts of the novel with a unified theme, different LLMs are used to continue the following text five times. Experts classified the composition of the five continuation results of the same model respectively, breaking down the continuation text into the three elements that make up the novel: environmental description, character development and storyline summary. Then bert-score is used to calculate the similarity among each element to preliminarily evaluate the homogeneity phenomenon of the novel text generated by LLMs. The experimental results are shown in Figure (b).}
\label{aex1}
\end{figure}

To verify homogenization phenomenon and identify the areas where the homogenization problem occurs most severely, we conducted an experiment on novel content generation based on the most popular Chinese LLMs, as shown in Figure~\ref{aex1}. Experimental results revealed that all LLMs exhibited a high degree of similarity in plot development, reinforcing the concern that homogenization reduces the diversity of literary works.

This phenomenon poses a threat to linguistic richness and narrative variety, weaken user's unique styles,  diminishing the creative value of user-generated content. Therefore, to solve the problem of homogeneity by LLMs, it is necessary to start from the generation logic of narrative texts by LLMs. Then it is possible to get rid of the same "three-part" hero story that is common in LLMs generation texts: "hero starts" - "hero conquers the enemy" - "hero triumphs".

\begin{table*}[t]
  \centering
  \resizebox{1\textwidth}{!}{
  \begin{tabular}{llll}
  \toprule
   Function (symbol) &Narrative function description & Function (symbol) &Narrative function description\\
   \midrule
    Initial situation (a) & The initial scene at the beginning of the story & Interdiction ($\gamma$) & A prohibition is imposed on the hero\\
    Interdiction violation ($\delta$) & The interdiction is violated & Reconnaissance ($\epsilon$) & The villain attempts to obtain information\\
    Delivery ($\xi$) & The villain gains information or an object & Trickery ($\eta$) & The villain deceives the hero to gain an advantage\\
    Complicity ($\theta$) & The victim submits to deception& Villainy (A) & The villain causes harm or injury\\
    Lack (a)& The scarcity of the protagonist & Mediation (B)& Misfortune is made known \\
    Counteraction (C)& The hero reacts to the villain’s actions & Departure ($\uparrow$) & The hero leaves home\\
    1st donor function (D) & The hero is tested by a potential donor& Hero’s reaction (E) & The hero reacts to the test\\
    Receipt of agent (F) &The hero acquires a magical agent & Transfer (G) &The hero is hero directed or taken to new location \\
    Struggle (H)& The hero and villain engage in direct conflict & Marking (J) & The hero is marked or identified \\
    Victory (I) & The hero defeats the villain. &Liquidation of lack (K) & Villainy or lack resolved \\
    Return and Pursuit ($\downarrow$)& The hero returns. The villain chases the hero& & \\
    Rescue (Rs) & The hero is saved from pursuit or danger &Unrecognized arrival (O) & The situation where the protagonist is not recognized \\
    Punishment (U)& The villain receives punishment or consequences& &either actively or passively\\
    Unfounded claims (L)& The protagonist is confronted with unreasonable & Transfiguration (T) & The hero undergoes a transformation.\\
    & demands, unfair competitions or difficulties &Wedding(W)  & The hero marries or is rewarded\\
    Solution (N)& The task is accomplished or resolved & Recognition (Q)& The hero is recognized or acknowledged \\
    Exposure (Ex) & The villain’s identity or deception is exposed & Difficult task (M) & The hero is given a difficult task\\
    
  \bottomrule
  \end{tabular}}
  \caption{Original Propp Narrative function list}
  \label{listf4}
\end{table*}

It is worth mentioning that our expert team consists of four people. Two people are mainly responsible for building the dataset and summarizing the list of new narrative functions. Among these two people, one is an online writer with over 5,000 + hours of writing experience, and the other is a seasoned online literature reader with over 10,000 hours. The remaining two are professionals with profound literary backgrounds and foundations. They are respectively a professor and a postgraduate student majoring in literature. They have extremely deep insights in the field of literature and are mainly responsible for the review of datasets and new lists to avoid errors.

\subsection{The list of original narrative functions.}Propp was a Russian formalist scholar in the early 20th century. His most significant contribution lies in the work "The Morphology of Stories" published in 1928. This magnum opus completely transformed people's understanding of folk tales and even the narrative structure itself, laying a solid foundation for modern narratology research and exerting a profound influence on later generations, such as structuralism, narratology, and the theory of story creation. Although each story character and plot detail he studied were different, these stories were essentially composed of a series of fixed and basic behavioral units. Propp referred to these basic behavioral units as "functions".  The list of the original thirty-one narrative functions mentioned in Propp's theory is shown in the Table~\ref{listf4}.

\subsection{Continue the text display.}
This section will present the continuation text based on Table 5. We respectively present the continuation texts of the top 2 rounds of all models to more intuitively reflect the plot paradigm. Tables~\ref{dp3} and \ref{dp2} respectively present the narrative sequences of "Lo-E-Em-P-S-U" and "Lo-H-E-Q-S-U" continued by Deepseek. Tables~\ref{qw1} and \ref{qw2} respectively show the narrative sequences of "A-Q-F-K-L-Fr-Lo" and "E-F-K-J-O-R-Re" that Qwen continued to write. Tables \ref{db1} and \ref{db2} respectively present the narrative sequence of "A-J-E-Lo-M-N-O" continued by Doubao.

\begin{table}[t]
	\centering
    \small
	\begin{tabularx}{0.5\textwidth}{X}
		\toprule
		   Narrative text generation\\
        \midrule
        \begin{CJK}{UTF8}{gbsn}
        苏业的大脑一片空白，他茫然地环顾四周，像个初生的婴儿一样无助（Lo）。黑衣人走近石台，低声自语：“只要能取出那枚封印之石，我便能掌控整个上古遗迹的力量，阻止我的野心的人都将付出代价——包括你这意外闯入者。”他手中升起一道黑光，轻触苏业的额头，一股阴冷的能量侵入，意图抹杀他残余的意识（H）。黑衣人的声音却暴露了线索——他似乎在寻找遗迹核心的钥匙（E）。苏业本能地颤抖，下一秒，他凭本能扑向黑衣人，拳头狠狠砸向对方（Q）。黑衣人轻蔑一笑，单掌一挥，黑光如鞭抽中苏业胸口，将他狠狠撞向石台；苏业瘫软倒地，浑身剧痛，意识濒临溃散（S）。他勉强爬起，捂伤逃跑，黑衣人紧随其后，火炬的热浪几乎灼烧他的背脊（U）。
        \end{CJK}\\
        \midrule
        Su Ye's mind went blank. He looked around blankly, as helpless as a newborn baby. The man in black approached the stone platform and muttered to himself in a low voice, "As long as I can take out that sealing stone, I will be able to control the power of the entire ancient ruins. Anyone who stops my ambition will pay the price - including you, the unexpected intruder." A black light rose from his hand, gently touching Su Ye's forehead. A cold energy invaded, intent on erasing his remaining consciousness (H). The voice of the man in black, however, revealed a clue - it seemed that he was seeking the key (E) to the core of the ruins. Su Ye instinctively trembled. In the next second, he instinctively pounced on the man in black and punched him hard (Q). The man in black gave a contemptuous smile, waved his single palm, and a black light like a whip struck Su Ye's chest, slamming him hard towards the stone platform. Su Ye collapsed to the ground, his body in excruciating pain, and his consciousness was on the verge of collapse. He struggled to his feet, covering his wounds and fleeing. The man in black followed closely behind, and the heat from the torch almost burned his back.\\
        \bottomrule
	\end{tabularx}
	\caption{The continuation text of Deepseek Ep.3}
    \label{dp3}
\end{table}%

\begin{table}[t]
	\centering
    \small
	\begin{tabularx}{0.5\textwidth}{X}
		\toprule
		   Narrative text generation\\
        \midrule
        \begin{CJK}{UTF8}{gbsn}
        苏叶的意识如同被漂白的布匹，只剩一片荒芜的空白（Lo）。他茫然地眨了眨眼，眼前这个从阴影中显现、英俊却陌生如刀刻的男人，胸腔里莫名翻搅着一种尖锐刺骨的痛感（E）。几乎在同一瞬间，一股狂潮般的酸楚与不知名的巨大愧怍决堤般冲垮了他的心房，随之而来的是无边无际的冰凉绝望，让他几乎窒息（Em）。
        “我不能…在这里…” 求生的念头压倒了那撕扯灵魂的情绪。趁着黑衣人起身未及完全警戒的刹那，苏叶像离弦的箭，用尽刚恢复的一丝气力朝与入口相反的、一条狭窄幽暗的岔隧道猛蹿进去！（P）。
        选择岔道并非完美。太窄了！苏叶的肩膀重重撞上嶙峋的岩壁，传来钻心的痛！虽然成功突入岔路几米，但他踉跄的身形暴露了虚弱与逃窜的狼狈，失败已在眼前（S）。
        “哼！” 身后传来一声冷厉不屑的轻嗤。黑衣人显然比他预想的更快。阴冷的寒意如跗骨之蛆骤然迫近，死亡的呼啸紧追着苏叶的后颈！他甚至能清晰感觉到追击者靴尖踢开碎石的风！在这黑暗狭窄的甬道里，一次呼吸的延迟就是永恒（U）。
        \end{CJK}\\
        \midrule
        Su Ye's consciousness was like a bleached piece of cloth, leaving only a desolate blank. He blinked blankly. The handsome yet strange man emerging from the shadows before him felt a sharp and bone-chilling pain churning in his chest for no reason. Almost at the same moment, a surging wave of sadness and an unknown sense of immense guilt burst through his heart like a dam, followed by an endless cold and desperate scene that almost suffocated him.
        I can't... Here..." The thought of survival overwhelmed the emotion that tore at the soul. Taking advantage of the moment when the man in black rose and was not yet fully alert, Su Ye, like an arrow released from a bow, used up the last bit of strength she had just regained and charged into a narrow and dark tunnel opposite the entrance! (P).
        Choosing a fork in the road is not perfect. It's too narrow! Su Ye's shoulder hit the craggy rock face hard, sending a piercing pain! Although he managed to break through a few meters into the fork in the road, his staggering figure exposed his weakness and the disheveled state of his escape. Failure was already within reach.
        "Humph!" " A cold and disdainful snort came from behind. The man in black was clearly faster than he had expected. The chilly chill, like a tarsal maggot, suddenly approached, and the roar of death closely followed Su Ye's back of the neck! He could even clearly feel the wind as the tips of the pursuers' boots kicked through the rubble! In this dark and narrow corridor, a delay in one breath is eternity (U).\\
        \bottomrule
	\end{tabularx}
	\caption{The continuation text of Deepseek Ep.2}
    \label{dp2}
\end{table}%

\begin{table}[t]
	\centering
    \small
	\begin{tabularx}{0.5\textwidth}{X}
		\toprule
		   Narrative text generation\\
        \midrule
        \begin{CJK}{UTF8}{gbsn}      
        苏业缓缓恢复意识，头痛得仿佛碎裂开来。模糊的视野中浮现出凹凸不平的洞壁，火炬残影已熄灭，只余晚光般微弱的磷光浮动（A）。转角阴影晃过，那蒙面黑衣人复现身寒毒视焦点灼灼锁定他心魂。“记忆消融反便利奴印！”嘶鸣狂飙挟带双掌如刃斩下；苏业神色骤缩混乱求生本能趋避侧滚腾挪开主轰路径，风刀呼啸擦额带血屑砸墙爆岩块（Q）。临近对斥缝隙中苏业瞥见黑衣人衣襟坠链浮动闪血琥珀核心雕刻诡秘图腾：密集三角嵌五芒繁丝—诡物象征着大规模灾祸策划之枢纽（F）。那刻苏阳苏醒于命弦绷张冰点，胸中滚溢破执生死畏缩岩浆般沸腾搏生志愿（K），他暴后登跃废石堆，旋风般穿过裂隧窄口遁向昏黑未知（L）。疾挺间腹部感奇内厚实暖意盘结流转循环微启—此为凝念驾驭大气精素根基，寻常修习者据此法采能可登元神开拓阶梯（Fr）。刚悟通毫末运转肌群纹期轨迹得携名册细解机要痕道谱系型格构造息忽，敌者嚎似索魂恶饕将银钉鞘设音波弹猛击中苏业顶骨中枢，再次封消满心握取记忆存拣新段位构成数据档案，拥半残意核坠沉入浑沌渊底永次忘失结构底页（Lo）。
        \end{CJK}\\
        \midrule
        Su Ye gradually regained consciousness, his headache aching as if it were shattering into pieces. In the blurred vision, uneven cave walls emerged. The remnants of the torch had gone out, leaving only A faint phosphorescent glow like evening light floating (A). At the corner, a shadow flashed by, and the masked man in black reappeared, his cold and poisonous gaze intently fixed on his soul. "Memory ablation is counterproductive!" " Neighing and surging, with both palms slashing down like blades; Su Ye's expression suddenly shrank and became chaotic. With the instinct for survival, he evaded and rolled sideways, opening up the main blasting path. The wind knife whistled, scraping his forehead with blood and debris, and smashed the wall and rock blocks. (Q) As Su Ye was approaching the crevice of the confrontation, he caught sight of the pendant chain on the black-clad man's robe floating and flashing blood. The core of the amber was engraved with a mysterious totem: dense triangles inlaid with five-pointed silk threads - the strange object symbolized the hub of large-scale disaster planning (F). At that moment, Su Yang awoke from the freezing point of his life string. His heart was filled with the determination to break free from the fear of life and death, boiling like lava to fight for survival (K). After the storm, he jumped onto the pile of abandoned rocks and, like a whirlwind, passed through the narrow opening of the tunnel and fled towards the dark and unknown (L). With a rapid rise, a thick and warm sensation is felt in the abdomen, swirling and circulating slightly. This is the foundation for concentrating the mind to control the essence of the atmosphere. Ordinary practitioners can use this method to ascend the ladder of primordial spirit development (Fr). Just after comprehending the trajectory of the muscle groups' lines, I was able to carry the list to carefully analyze the key structure of the trace path lineage. Suddenly, the enemy's howl was like a vengeful soul, and the silver nail sheath was set up with a sound wave bullet to strike the center of the middle Suye's parietic bone. Once again, I sealed my heart and held the memory to store a new position to form a data file. With a half-broken mind core, I sank into the bottom of the chaos abyss and lost the structure of the bottom page (Lo).\\
        \bottomrule
	\end{tabularx}
	\caption{The continuation text of Qwen Ep.1}
    \label{qw1}
\end{table}%

\begin{table}[t]
	\centering
    \small
	\begin{tabularx}{0.5\textwidth}{X}
		\toprule
		   Narrative text generation\\
        \midrule
        \begin{CJK}{UTF8}{gbsn}      
        黑衣人凝视他失神之态，陡然冷笑展开掌心幻影，呈现扭曲光幕间闪烁的裂隙坐标：“东南角通幽裂隙七分后闭合...此变异动源于噬魔外源镇压释放!”光点如血渗入苏业眼底，映射出崩碎禁地的轨迹(E) 。叙毕，黑衣人踉跄掩胸咳嗽时喉中混响异样呻吟:“若坠日侵染九层境的主核再难覆敛...届时满愿表柱的谜题必启亡荒世纪”;语末甩袖抛落半焦紫符坠入地缝，引起空间涟漪暴动(F) 。苏业直觉言语如刃割裂理性，却强抑恐慌喉头蠕动心生新念:必须截断这诡异的侵袭循环!(K) 恰此时脚边火炬微光映照石壁龟裂处泛起青纹符语(J):‘溯源之眼融祭生于天枢坤托’。字痕灼热间袖袋突凸物刺动神经，他疾手掏出—竟是刻日月双环嵌合的古币温转于指，墨蓝电光覆脑刹那感知外界流速，空间位移能力激增(O)。币体分溶解鳞状纹路缠右腕成形流金三角徽印，宛若星链呼唤起源证记烙印不散(R)。“卍解劫环灭叠持有躯壳!初始伪消元铸须脉世系的连锁延续者吧!”尖嚎骤起—黑衣人面皮溶落露骨相眉弧刺亮，惊现苏业幼年梦痕刻存的其母容颜!血泪滴露化万千意识洪流席卷：“你终是墟灭废墟唯一遗骨承载人...诸界仅认血脉醒证者！”哭啸交融撼透回路认知紧锁(Re)。
        \end{CJK}\\
        \midrule
        The man in black stared at his absent-minded expression, suddenly sneering and unfolding the phantom in his palm, revealing the coordinates of the fissure flickering between the distorted light screens: "The fissure at the southeast corner will close after seven minutes..." This mutation movement originated from the external suppression release of the Goblin!" The light spot, like blood, seeped into Su Ye's eyes, reflecting the trajectory of the shattered forbidden area (E). After the narration, the man in black staggered, covered his chest and coughed. A strange moan reverberated in his throat: "If the setting sun infects the main core of the Ninth Layer Realm and can no longer be contained..." At that time, the mystery of the fulfillment pillar will surely usher in a desolate era. At the end of the speech, he flung his sleeve and dropped a half-burnt purple talisman, which fell into a crack in the ground, causing a ripple riot in the space (F). Su Ye felt that his words were like a sharp blade cutting through reason, but he managed to suppress his panic. A new thought emerged in his throat: This strange cycle of invasion must be cut off! (K) Just at this moment, the faint light of the torch at the feet shone on the cracked part of the stone wall, causing a bluish-patterned symbol (J) to appear: "The eye that traces back to the source, the sacrificial being born in the celestial Shu Kuntuo." The burning marks on his sleeve bag pricked his nerves. He quickly pulled it out - it turned out to be an ancient coin with two rings engraved with the sun and the moon, warm and rotating in his fingers. The dark blue electric light covered his brain, and in an instant, he sensed the flow rate of the outside world, and his spatial displacement ability increased sharply (O). The coin body is composed of dissolved scale-like patterns wrapped around the right wrist to form a golden triangular emblem, resembling a starry chain calling for an indelible origin mark (R). "Bankai Jie Huan Mie Jie Possesses a body!" "Be the chain continuation of the initial pseudo-Elimination Yuan Casting Xu lineage!" " A sudden shrill howl rang out - the black-clad man's face melted, revealing his bony features, and his eyebrows shone brightly, revealing the face of his mother that Su Ye had dreamed of in his childhood! Blood and tears, drops of dew, transform into a torrent of consciousness sweeping across: "You are ultimately the only one to bear the remains of the ruins that have been destroyed..." All realms only recognize those who are awakened by the bloodline!" The interplay of crying and screaming shatters the cognitive lock of the loop (Re).\\
        \bottomrule
	\end{tabularx}
	\caption{The continuation text of Qwen Ep.2}
    \label{qw2}
\end{table}%

\begin{table}[t]
	\centering
    \small
	\begin{tabularx}{0.5\textwidth}{X}
		\toprule
		   Narrative text generation\\
        \midrule
        \begin{CJK}{UTF8}{gbsn}      
        苏叶在刺骨的寒意中睁眼，火炬残焰在石壁上投下摇曳的光影，潮湿的空气里混杂着泥土与铁锈的气息，石台边缘还凝着未化的水珠（A）。
        身前的黑衣人已收起斗篷，见他苏醒，沉声道：“我叫陵光，是你父亲的护卫。你本是苍岚宗少主，方才你看到的是我体内的‘护主印’，与你血脉共鸣引发能量波动，才让你失去记忆”（J）。
        苏叶瞳孔微缩，指尖下意识攥紧身下的石面，“父亲”“苍岚宗” 这些词汇像碎片般撞入脑海，却无法拼凑完整，心底只剩莫名的悸动与茫然（E）。他用力皱眉回想，脑海依旧一片空白，连自己的名字都需反复确认（Lo）。
        陵光抬手，掌心浮现一枚莹白玉佩，玉佩流转的光晕与苏叶体内微弱的气流隐隐呼应，显然是同源的灵力波动（M）。
        苏叶望着那枚玉佩，警惕的神色渐渐缓和，迟疑着伸出手，指尖刚要触碰到玉佩，又微微停顿，眼神里满是不确定的探寻（N）。
        陵光将玉佩放入他掌心，玉佩瞬间融入指尖，一股暖流顺着经脉蔓延全身，原本虚弱的体力尽数恢复，丹田处还浮现出淡淡的灵力光晕，他甚至能清晰感知到周围气流的流动（O）。
        \end{CJK}\\
        \midrule
        Su Ye opened her eyes in the bone-chilling cold. The remaining flame of the torch cast flickering shadows on the stone wall. The damp air was mixed with the smell of earth and rust, and there were still unmelted water droplets (A) condensing at the edge of the stone platform.
        The man in black in front of him had already put away his cloak. Seeing him wake up, he said in a deep voice, "My name is Lingguang. I'm your father's guard." You were originally the young master of the Canglan Sect. Just now, what you saw was the "Guardian Seal" within me. It resonated with your bloodline, causing energy fluctuations, which led to your loss of memory. (J)   
        Su Ye's pupils shrank slightly, and her fingertips subconsciously tightened the stone surface beneath. Words like "father" and "Canglan Sect" smashed into her mind like fragments, but she couldn't piece them together completely. All that remained in her heart was an inexplicable fluttering and confusion. He frowned forcefully and recalled, his mind still blank, even needing to repeatedly confirm his own name. Lingguang raised his hand, and a lustrous white jade pendant emerged in the palm of his hand. The halo flowing from the pendant faintly echoed with the faint airflow within Su Ye's body, clearly indicating that they were the same source of spiritual energy fluctuations (M).   
        Su Ye gazed at the jade pendant. Her vigilant expression gradually eased. Hesitantly, she reached out her hand. Just as her fingertips were about to touch the pendant, she paused slightly, her eyes filled with an uncertain search(N).
        Lingguang placed the jade pendant in the palm of his hand. Instantly, the pendant merged into his fingertips. A warm current spread throughout his body along the meridians. His originally weak physical strength was fully restored, and a faint aura of spiritual energy emerged in his dantian. He could even clearly sense the flow of the surrounding air currents(O).\\
        \bottomrule
	\end{tabularx}
	\caption{The continuation text of Doubao Ep.1}
    \label{db1}
\end{table}%

\begin{table}[t]
	\centering
    \small
	\begin{tabularx}{0.5\textwidth}{X}
		\toprule
		   Narrative text generation\\
        \midrule
        \begin{CJK}{UTF8}{gbsn}      
        苏叶在昏沉中睁眼，火炬的光晕在狭窄裂缝中铺展，岩壁渗出的水汽打湿了他的衣角，石台冰凉的触感透过衣衫传来，空气中飘着淡淡的石屑与草木混合的气息（A）。身前的黑衣人静静伫立，见他苏醒，开口道：“我叫砚辞，是守护你家族的暗卫。你是灵溪族最后的血脉，方才我卸下伪装时，体内的族印与你血脉共振，引发能量冲击让你失了忆”（J）。苏叶喉结滚动，“灵溪族”“族印” 这些陌生的词汇让他心头一紧，指尖不自觉蜷缩，眼神里满是困惑与警惕，仿佛想从对方脸上找到答案（E）。他用力闭眼回想，脑海中只有一片混沌，连自己的过往都成了模糊的空白，甚至不确定 “苏叶” 这个名字是否属于自己（Lo）。砚辞抬手，掌心浮现一枚刻着藤蔓纹路的青铜令牌，令牌上流转的微光与苏叶体内微弱的悸动隐隐呼应，显然是同源的血脉力量（M）。苏叶盯着青铜令牌，警惕的神色稍稍松动，迟疑着往前挪了半步，伸出的手在半空微微停顿，眼神里藏着不确定的探寻（N）。砚辞将令牌递到他掌心，令牌触肤的瞬间骤然发热，一股温暖的能量顺着手臂涌入全身，原本虚弱的体力瞬间充盈，丹田处还浮现出淡淡的绿光，他能清晰感知到周围草木的气息，指尖甚至能牵引一缕微弱的草木灵力（O）。
        \end{CJK}\\
        \midrule
        Su Ye opened his eyes in A daze. The halo of the torch spread through the narrow crack. The water vapor seeping from the rock wall moistened the corner of his clothes. The cold touch of the stone platform came through his clothes. A faint smell of a mixture of stone chips and grass and trees was floating in the air(A).
        The black-clad man in front of him stood still. Seeing him wake up, he said, "My name is Yan Ci. I'm the secret guard protecting your family." You are the last bloodline of the Lingxi clan. Just now, when I took off my disguise, the clan seal within my body resonated with your bloodline, triggering an energy shock that made you lose your memory. (J)
        Su Ye's Adam's apple rolled. The unfamiliar words like "Lingxi Tribe" and "Tribe Seal" made his heart tighten(E). His fingertips involuntarily curled up, and his eyes were filled with confusion and vigilance, as if he wanted to find the answer on the other person's face. He closed his eyes forcefully and recalled. In his mind was nothing but chaos. Even his past had become a vague blank, and he wasn't even sure if the name "Su Ye" belonged to him (Lo).   
        Yan Ci raised her hand, and a bronze token engraved with vine patterns emerged in the palm of her hand. The faint light flowing on the token faintly echoed the faint palpitations within Su Ye's body, clearly representing the same source of bloodline power (M).
        Su Ye stared at the bronze token, his vigilant expression slightly relaxed. Hesitantly, he moved forward half a step, his outstretched hand pausing slightly in mid-air, with an uncertain search hidden in his eyes(N).
        Yan Ci handed the token to his palm. The moment the token touched his skin, it suddenly heated up, and a warm energy rushed into his entire body through his arm. His originally weak physical strength instantly filled up, and a faint green light emerged in his dantian. He could clearly sense the scent of the surrounding grass and trees, and his fingertips could even draw a faint wisp of grass and tree spiritual energy (O).\\
        \bottomrule
	\end{tabularx}
	\caption{The continuation text of Doubao Ep.2}
    \label{db2}
\end{table}%

\subsection{Analysis of the Functional Paradigm of Plot Narrative}
In the experiment of analyzing the narrative paradigms of storylines, we summarized six narrative arrangement paradigms to represent six different plots. We present in detail the text narrative functions obtained from sampling in Table~\ref{caseSam} to more clearly explain how we analyzed and derived the six paradigms. We first listed the samples that conformed to the summary paradigm, and then listed the samples that did not conform to the paradigm.

\begin{table*}[t]
  \centering
  \small
  \setlength{\tabcolsep}{0.5mm}
  \resizebox{\linewidth}{!}{
  \begin{tabular}{lllllll}
  \toprule
   No. & Battle & Emotional & Difficult task & Adventure & Pretending & Daily life \\
   \midrule
    1 & A-F-H-K-Q-S & Em-A-K-E-Ch & Y-F-Em-K-Q-A-H-Z & P-H-Em-K-O & W-De-A-Q-S & A-K-F-E-Ch \\
    2 & A-D-E-K-Q-O & Em-K-F-E-Q-Ch & Y-A-Q-I-H-Em-F-K-Z & P-A-Q-F-I-M & W-K-De-Em-S & A-E-Em-J-H-Ch \\
    3 & A-F-I-K-L-Q-S & Em-E-A-K-O-Ch & Y-Em-A-F-K-Q-H-I-Z & P-E-De-K-H-L-O & W-F-B-De-S & A-K-J-F-Em-Ch \\
    4 & A-E-H-K-Q-O & Em-K-F-Re-L-Ch & Y-K-Q-F-Em-H-A-E-Z & P-Q-Re-F-H-A-K-O & W-De-E-Ch-H-S & A-E-K-F-I-Ch \\
    5 & A-D-F-K-Q-S & Em-F-E-K-Fi-Ch & Y-E-K-H-A-Q-F-Em-Z & P-Em-F-I-B-C-G-E-M & W-I-De-C-Q-S & A-Em-H-F-K-Ch \\
    6 & A-F-Em-K-Q-O & Em-K-F-Q-De-Ch & Y-I-Em-Q-K-F-H-A-Z & P-K-H-F-O & W-A-K-De-F-S & A-J-K-Em-I-Ch \\
    7 & A-E-I-K-Q-S & Em-E-K-A-R-Ch & Y-H-A-Em-I-K-Q-F-Z & P-A-E-Q-Fi-H-K-L-X-M & W-Em-H-D-De-S & A-F-E-K-Em-Ch \\
    8 & A-D-H-K-Q-O & Em-F-K-E-W-Ch & Y-Q-Em-K-F-H-I-E-Z & P-I-F-De-Em-O & W-Q-De-I-E-Ch-S & A-I-J-H-E-Ch \\
    9 & A-F-K-L-Q-S & Em-K-E-F-X-Ch & Y-F-K-Em-Q-I-H-A-Z & P-H-K-E-Fa-I-M & W-G-F-De-H-I-S & A-K-Em-F-J-Ch \\
    10 & A-E-Em-K-Q-O & Em-F-E-K-Y-Ch & Y-A-H-K-I-Em-F-Q-Z & P-Q-Em-F-Re-H-A-K-Lo-E-O & W-De-K-A-L-Em-S & A-E-K-K-Em-Ch \\
    11 & A-D-I-K-Q-S & Em-K-A-E-Z-Ch & Y-Em-K-Q-F-H-A-E-Z & P-F-E-A-Q-M & W-I-F-Q-De-Ch-S & A-H-I-F-K-Ch \\
    12 & A-F-H-K-L-Q-O & Em-E-F-K-Fa-Ch & Y-K-I-Q-Em-H-F-A-Z & P-Em-Q-R-Fi-H-O & W-De-H-E-Em-K-O-S & A-Em-J-K-I-Ch \\
    13 & A-E-K-P-Q-S & Em-F-K-E-G-Ch & Y-E-Q-A-K-F-Em-H-Z & P-G-H-I-F-M & W-A-De-I-Ch-F-Q-S & A-F-K-E-J-Ch \\
    14 & A-D-F-K-Q-O & Em-K-E-K-V-Ch & Y-I-H-F-K-Em-Q-A-Z & P-A-E-K-Fr-L-M & W-K-H-De-E-R-I-S & A-K-H-Em-F-Ch \\
    15 & A-F-I-K-Q-S & Em-E-K-F-K-Ch & Y-H-Em-F-I-Q-K-E-Z & P-Q-Fa-H-K-Em-O & W-F-Q-I-De-Ch-Em-S & A-J-Em-F-H-Ch \\
    16 & A-E-H-K-L-Q-O & Em-F-K-E-Lo-Ch & Y-Q-F-H-A-Em-I-K-Z & P-E-I-K-L-M & W-De-E-H-A-K-Q-I-S & A-E-F-K-H-Ch \\
    17 & A-D-K-P-Q-S & Em-K-F-E-O-Ch & Y-F-A-Q-Em-K-H-I-Z & P-B-A-Em-F-O & W-Em-De-Ch-F-I-M-H-S & A-K-I-J-Em-Ch \\
    18 & A-F-Em-K-Q-S & Em-E-F-K-J-Ch & Y-A-K-H-F-Q-Em-I-Z & P-H-I-C-Em-J-O & W-Ch-De-I-S & A-F-J-E-K-Ch \\
    19 & A-E-I-K-Q-O & Em-A-J-K-N-Ch & Y-Em-H-Q-I-F-K-A-Z & P-K-Em-F-G-H-M & W-De-E-Em-S & A-H-K-Em-J-Ch \\
    20 & A-D-H-K-P-Q-S & Em-K-F-E-M-Ch & Y-K-F-I-A-H-Em-Q-Z & P-A-Q-E-K-N-O & W-H-K-De-F-S & A-E-Em-F-I-Ch \\
    21 & A-F-K-Q-R-O & Em-E-K-F-Fr-Ch & Y-E-Em-H-A-I-F-K-Z & P-Em-H-F-I-P-M & W-H-De-K-Em-A-S & A-K-F-J-I-Ch \\
    22 & A-E-Em-K-Q-S & Em-F-K-E-U-Ch & Y-I-Q-K-Em-A-H-F-Z & P-F-D-K-H-Em-O & W-K-Em-De-I-S & A-J-F-K-Em-Ch \\
    23 & A-D-I-K-Q-O & Em-K-F-E-A-Ch & Y-H-F-K-I-A-Em-Q-Z & P-E-Q-Fa-H-I-K-M & W-Q-F-E-De-Ch-S & A-Em-K-J-H-Ch \\
    24 & A-F-H-K-Q-R-S & Em-E-F-K-W-Ch & Y-Q-I-H-K-F-A-Em-Z & P-K-F-Em-Lo-Q-S-O & W-De-I-H-A-N-S & A-F-H-Em-K-Ch \\
    25 & A-E-K-Q-O & Em-F-K-E-De-Ch & Y-F-K-A-H-Em-I-Q-Z & P-A-H-Em-F-V-K-M & W-K-De-Em-Ch-F-Q-S & A-E-J-Em-F-Ch \\
    26 & A-D-F-K-P-Q-S & Em-K-E-F-Z-Fi-Ch & Y-A-Em-Q-H-I-F-K-Z & P-Em-F-G-H-I-K-O & W-Em-H-De-F-Q-I-S & A-K-Em-I-F-Ch \\
    27 & A-F-I-K-Q-R-O & Em-E-K-F-X-Q-Ch & Y-Em-F-H-Q-I-K-A-Z & P-Q-I-F-H-Em-T-M & W-A-Q-De-H-K-S & A-H-F-E-K-Ch \\
    28 & A-E-H-K-Q-S & Em-F-K-E-Y-S-Ch & Y-K-I-Em-A-Q-H-F-Z & P-K-E-Fa-R-Em-H-O & W-De-Ch-K-Em-F-S & A-J-I-K-F-Ch \\
    29 & A-D-K-Q-O & Em-K-F-E-R-Fi-Ch & Y-E-H-Q-K-A-F-Em-Z & P-H-Em-I-F-W-K-O & W-I-K-H-De-Em-S & A-K-E-J-Em-Ch \\
    30 & A-F-Em-K-Q-R-S & Em-E-F-K-E-Re-Ch & Y-I-F-A-K-Q-Em-H-Z & P-A-Q-K-Fi-Em-L-M & W-F-E-De-Q-K-S & A-F-K-H-I-Ch \\
    31 & A-E-I-K-Q-O & Em-F-K-E-P-A-Ch & Y-H-Q-F-Em-I-K-A-Z & P-E-H-I-K-F-Z-O & W-De-Em-H-I-Q-S & A-Em-F-E-K-Ch \\
    32 & A-D-H-K-Q-S & Em-K-E-F-Q-Re-Ch & Y-Q-A-E-H-F-K-I-Z & P-Em-A-K-F-De-I-M & W-K-F-De-I-Em-S & A-E-H-J-K-Ch \\
    33 & A-F-K-Q-O & Em-E-K-F-K-Fi-Ch & Y-A-H-F-Em-E-Q-K-Z & P-F-Q-H-Em-J-K-O & W-H-I-De-F-Em-S & A-K-J-Em-F-Ch \\
    34 & A-E-Em-K-Q-R-O & Em-F-K-E-V-S-Ch & Y-Em-I-K-F-Q-A-H-Z & P-K-Em-F-I-Lo-P-Q-M & W-De-A-Em-K-F-S & A-I-Em-E-F-Ch \\
    35 & A-D-I-K-Q-S & Em-K-F-E-Lo-R-Ch & Y-K-F-H-Em-Q-I-A-Z & P-H-A-Em-F-K-R-O & W-Ch-Q-De-I-F-S & A-F-H-K-J-Ch \\
    36 & A-F-H-K-Q-O & Em-E-F-K-M-De-Ch & Y-E-Q-K-I-Em-F-H-Z & P-Q-E-Em-H-T-U-I-O & W-Em-I-De-K-H-S & A-E-K-Em-H-Ch \\
    37 & A-E-K-Q-S & Em-F-K-E-J-Q-Ch & Y-I-Em-H-F-A-K-Q-Z & R-A-Q-F-Em-H-S & W-F-K-De-H-Q-S & A-J-K-F-Em-Ch \\
    \cline{5-5}
    38 & A-D-F-K-Q-O & Em-K-E-F-O-Re-Ch & Y-H-K-Q-F-I-Em-A-Z & Fi-E-K-H-I-Lo & W-Q-K-De-Em-I-S & A-K-F-Em-I-Ch \\
    39 & A-F-I-K-Q-S & Em-E-K-F-Fa-A-Ch & Y-Q-Em-A-I-H-K-F-Z & Em-H-I-J-K-F & W-H-Em-De-A-F-S & A-H-Em-K-E-Ch \\
    40 & A-E-H-K-Q-O & Em-F-K-E-N-Fi-Ch & Y-F-Q-K-Em-H-A-E-Z & F-G-K-Em-H-Z & W-I-Em-De-Q-K-S & A-F-I-E-Em-Ch \\
    \cline{2-2}
    41 & A-E-H-Q & Em-K-F-E-Fr-E-Ch & Y-A-I-Em-F-Q-K-H-Z & K-H-Em-I-Fr & W-De-Lo-A-K-S & A-E-K-F-Em-H-Ch \\
    42 & E-F-K-Q-O & Em-E-F-K-A-Re-Ch & Y-Em-H-F-K-I-Q-A-Z & Q-Em-F-A-B-C-H & W-Fr-Em-De-I-H-S & A-K-Em-J-F-I-Ch \\
    43 & A-C-D-K & Em-F-K-E-G-Fa-Ch & Y-K-A-F-I-Em-H-Q-Z & Em-F-G-H-I-L-M & A-Q-K-F-Em & A-J-E-H-K-F-Ch \\
    \cline{3-3}
    \cline{6-6}
    44 & K-Fr-Q-S-Fi-Z & K-F-E-Q-S & Y-E-I-Q-F-H-K-Em-Z & De-A-Q-F-Em-H-O & Ch-H-I-Em-Q & A-F-K-Em-E-J-Ch \\
    45 & E-K-P-R-Re-Lo & E-K-F-Re-Fa & Y-H-F-A-Q-Em-I-K-Z & A-K-Em-F-I-P & De-Fa-K-Em-I & A-I-K-H-F-Em-Ch \\
    46 & F-Y-Fa-L-Q-Re & F-E-K-X-Y & Y-Q-K-Em-I-F-A-H-Z & Fr-E-Em-H-I-J & Lo-Fr-A-Q-K & A-E-Em-K-I-H-Ch \\
    47 & O-K-M-Q-Re-Fi & K-F-E-Z-Fi & Y-F-H-I-Q-A-K-Em-Z & H-Em-Q-F-K-L & I-H-Em-K-Ch-F & A-K-F-J-E-Em-Ch \\
    48 & X-K-Q-S-R-De & E-K-F-De & Y-A-E-H-Em-Q-F-K-Z & I-F-Em-H-Re-M & B-A-Q-I-Em-Ch & A-H-J-Em-K-F-Ch \\
    49 & L-E-Fr-Q-S-Fa & F-K-E-W & Y-Em-K-H-F-A-I-Q-Z & Lo-K-Em-F-G-H & E-K-F-Q-H-I & A-Em-E-K-J-F-Ch \\
    50 & J-Y-Q-S-Fi-W & K-E-F-Q & Y-F-Em-Y-Q-K-H-A-Z & Re-A-Em-F-H-K & Fa-Em-Ch-H-K & A-F-I-K-E-H-Ch \\
    51 & F-K-P-O-Q-Re & F-K-E-R & Y-Lo-B-Fr-Y-A-H-Z & Ch-Q-K-Em-F-I & Q-I-K-H-Em-F & A-K-H-F-Em-J-Ch \\
    \cline{4-4}
    52 & M-E-K-Q-S-V & E-K-F-Lo & H-I-A-Em-K-P-R & Fa-H-I-Em-F-K & Re-De-F-E-M & A-E-J-F-K-H-Ch \\
    53 & P-K-Q-R-S-De & K-F-E-G & D-L-F-J-Q-Em-X & A-Em-Q-F-H-O & G-Em-I-K-F-Q & A-Em-I-H-J-K-Ch \\
    \cline{7-7}
    54 & Lo-K-F-Q-S-L & F-E-K-P & B-K-Em-F-Q-Fr-N & F-Em-R-S-T-U & Lo-Fi-H-K-I & G-Fr-De-Lo-B \\
    55 & N-K-Q-Re-Y & E-K-F-V & V-M-Q-Em-H-S-O & Em-H-I-Fa-K-L & A-Em-Ch-H-I-K & V-Em-I-K-F-P \\
    56 & Q-S-Fi-M-Fa & F-K-E-N & Lo-F-C-A-Em-Z-K & De-F-Em-H-Q & Q-K-F-I-Em-H & T-H-Q-Em-J-R \\
    57 & E-K-P-Q-S-Fa & K-E-F-M & S-W-B-Em-L-H-K & G-K-Em-H-I & Fi-Lo-A-Q-K & S-Fi-Re-Em-K-W \\
    58 & J-K-Q-S-Fr & F-E-K-U & G-Re-Fa-Q-F-Em & Lo-Em-H-I-K & Ch-Em-H-K-I-F & U-O-F-Lo-H-Z \\
    59 & F-K-Q-Re-W & E-K-F-O & A-Fr-Y-K-Z-Ch-D & Fi-F-Em-H-A & D-E-Em-F-Q-I & N-K-E-K-Em-V \\
    60 & P-K-Q-Fi-R & K-F-E-J & M-J-Em-H-K-Y & R-E-Em-K-I-S & Fr-Re-K-Em-A & P-I-Fa-F-Em-C \\
  \bottomrule
  \end{tabular}}
  \caption{Sampling analysis of the functional paradigm of plot narrative}
  \label{caseSam}
\end{table*}

\subsection{A complete list of online literature books}
We have placed a detailed list of 100 online literature novels collected by the National Library of China in Tables~\ref{100part1} and Tables~\ref{100part2}, including the title, topic and word count of each book. For more detailed information, please refer to \textit{http://dportal.nlc.cn:8332/zylb/zylb.htm}.

\begin{table*}[!htbp]
	\centering
    \small
	\begin{tabularx}{1\textwidth}{l|l|l}
		\toprule
		 Name &  Topic & Words \\
        \midrule
        \begin{CJK}{UTF8}{gbsn}诡秘之主\end{CJK}/Lord of Mysteries & Fantasy & 4.47M \\
        \begin{CJK}{UTF8}{gbsn}武道宗师\end{CJK}/Martial Arts Grandmaster & Fantasy & 2.48M\\
        \begin{CJK}{UTF8}{gbsn}斗破苍穹\end{CJK}/Battle Through the Heavens & Fantasy & 5.33M\\
        \begin{CJK}{UTF8}{gbsn}斗罗大陆\end{CJK}/Douluo Dalu & Fantasy & 2.97M\\
        \begin{CJK}{UTF8}{gbsn}大道朝天\end{CJK}/The Path Towards Heaven & Fantasy & 3.00M\\
        \begin{CJK}{UTF8}{gbsn}武极天下\end{CJK}/Martial World & Fantasy & 6.66M\\
        \begin{CJK}{UTF8}{gbsn}圣墟\end{CJK}/Sacred Ruins & Fantasy & 5.64M\\
        \begin{CJK}{UTF8}{gbsn}天域苍穹\end{CJK}/Sky Soaring & Fantasy & 5.05M\\
        \begin{CJK}{UTF8}{gbsn}太古神王\end{CJK}/Ancient God King & Fantasy & 6.32M\\
        \begin{CJK}{UTF8}{gbsn}气冲星河\end{CJK}/Spirit Impacting Galaxies & Fantasy & 3.24M\\
        \begin{CJK}{UTF8}{gbsn}光明纪元\end{CJK}/Era of Illumination & Fantasy & 7.00M\\
        \begin{CJK}{UTF8}{gbsn}巫神纪\end{CJK}/Record of the Witch God & Fantasy & 4.11M\\
        \begin{CJK}{UTF8}{gbsn}牧神记\end{CJK}/Pastoral God Memoirs & Fantasy & 5.89M\\
        \begin{CJK}{UTF8}{gbsn}十州风云志\end{CJK}/Ten States Chronicles & Fantasy & 1.28M\\
        \begin{CJK}{UTF8}{gbsn}全职法师\end{CJK}/Full-Time Magister & Fantasy & 7.08M\\
        \begin{CJK}{UTF8}{gbsn}弑天刃\end{CJK}/Sky Reaping Blade & Fantasy & 5.18M\\
        \begin{CJK}{UTF8}{gbsn}太上章\end{CJK}/Venerable Code & Fantasy & 3.95M\\
        \begin{CJK}{UTF8}{gbsn}异常生物见闻录\end{CJK}/Records of Unusual Creatures & Fantasy & 5.25M\\
        \begin{CJK}{UTF8}{gbsn}死在火星上\end{CJK}/Death on Mars & Fantasy & 5.9M\\
        \begin{CJK}{UTF8}{gbsn}修真四万年\end{CJK}/Cultivation for Forty Millennia & Fantasy & 10.80M\\
        \begin{CJK}{UTF8}{gbsn}地球纪元\end{CJK}/Earth Era & Fantasy & 1.42M\\
        \begin{CJK}{UTF8}{gbsn}盘龙\end{CJK}/Coiling Dragon & Fantasy & 3.33M\\
        \begin{CJK}{UTF8}{gbsn}放开那个女巫\end{CJK}/Release That Witch & Fantasy & 3.37M\\
        \begin{CJK}{UTF8}{gbsn}全职高手\end{CJK}/The King's Avatar & Fantasy & 5.35M\\
        \begin{CJK}{UTF8}{gbsn}惊悚乐园\end{CJK}/Thrill Paradise & Fantasy & 4.89M\\
        \begin{CJK}{UTF8}{gbsn}诛仙\end{CJK}/Renegade Immortal & Xianxia & 1.51M\\
        \begin{CJK}{UTF8}{gbsn}从前有座灵剑山\end{CJK}/Once Upon a Time on Spirit Sword Mountain & Xianxia & 2.88M\\
        \begin{CJK}{UTF8}{gbsn}不朽凡人\end{CJK}/Immortal Mortal & Xianxia & 3.69M\\
        \begin{CJK}{UTF8}{gbsn}一念永恒\end{CJK}/A Thought Through Eternity & Xianxia & 3.69M\\
        \begin{CJK}{UTF8}{gbsn}斗战狂潮\end{CJK}/Battle Mania Tide & Xianxia & 4.10M\\
        \begin{CJK}{UTF8}{gbsn}仙葫\end{CJK}/Gourd Immortal & Xianxia & 2.56M\\
        \begin{CJK}{UTF8}{gbsn}凡人修仙传\end{CJK}/A Record of a Mortal's Journey to Immortality & Xianxia & 7.48M\\
        \begin{CJK}{UTF8}{gbsn}星辰变\end{CJK}/Stellar Transformations & Xianxia & 2.81M\\
        \begin{CJK}{UTF8}{gbsn}道君\end{CJK}/Dao Lord & Xianxia & 5.01M\\
        \begin{CJK}{UTF8}{gbsn}庆余年\end{CJK}/Celebrating Years Beyond & Time travel & 3.78M\\
        \begin{CJK}{UTF8}{gbsn}赘婿\end{CJK}/The Unwanted Son-in-Law & Time travel & 4.86M\\
        \begin{CJK}{UTF8}{gbsn}孺子帝\end{CJK}/The Child Emperor & Time travel & 1.80M\\
        \begin{CJK}{UTF8}{gbsn}朱门风流\end{CJK}/Scholarship among Vermilion Doors & Time travel & 3.33M\\
        \begin{CJK}{UTF8}{gbsn}宋时行\end{CJK}/Journey Through the Song Dynasty & Time travel & 1.57M\\
        \begin{CJK}{UTF8}{gbsn}唐砖\end{CJK}/Tang Dynasty Brick & Time travel & 4.47M\\
        \begin{CJK}{UTF8}{gbsn}覆汉\end{CJK}/Overthrowing Han & Time travel & 3.23M\\
        \begin{CJK}{UTF8}{gbsn}秦吏\end{CJK}/Qin Bureaucrat & Time travel & 3.36M\\
        \begin{CJK}{UTF8}{gbsn}官居一品\end{CJK}/A First-Rank Official & Time travel & 5.71M\\
        \begin{CJK}{UTF8}{gbsn}医统江山\end{CJK}/Healing the Nation's Rivers and Mountains & Time travel & 5.24M\\
        \begin{CJK}{UTF8}{gbsn}明朝败家子\end{CJK}/The Spendthrift of the Ming Dynasty & Time travel & 5.42M\\
        \begin{CJK}{UTF8}{gbsn}大明望族\end{CJK}/Prestigious Family of the Ming & Time travel & 3.08M\\
        \begin{CJK}{UTF8}{gbsn}上品寒士\end{CJK}/Refined Humble Scholar & Time travel & 1.51M\\
        \begin{CJK}{UTF8}{gbsn}贞观大闲人\end{CJK}/Idle Man of Zhenguan & Time travel & 3.28M\\
        \begin{CJK}{UTF8}{gbsn}民国谍影\end{CJK}/Shadows of the Republic & Time travel & 3.72M\\
        \begin{CJK}{UTF8}{gbsn}冠军之心\end{CJK}/The Heart of a Champion & Time travel & 3.68M\\

         \bottomrule
	\end{tabularx}
	\caption{100 online novels collected by the National Library of China (part1)}
    \label{100part1}
\end{table*}%

\begin{table*}[!htbp]
	\centering
    \small
	\begin{tabularx}{1\textwidth}{l|l|l}
		\toprule
		 Name &  Topic & Words \\
        \midrule
        \begin{CJK}{UTF8}{gbsn}复兴之路\end{CJK}/Path to Revitalization & Time travel & 9.6M\\
        \begin{CJK}{UTF8}{gbsn}朝阳警事\end{CJK}/Chaoyang Police Stories & City & 2.10M\\
        \begin{CJK}{UTF8}{gbsn}重燃\end{CJK}/Reigniting & City & 2.28M\\
        \begin{CJK}{UTF8}{gbsn}回到过去变成猫\end{CJK}/Back to the Past as a Cat & City & 1.44M\\
        \begin{CJK}{UTF8}{gbsn}神藏\end{CJK}/Divine Collection & City & 5.22M\\
        \begin{CJK}{UTF8}{gbsn}美食供应商\end{CJK}/Gourmet Supplier & City & 5.29M\\
        \begin{CJK}{UTF8}{gbsn}宠物天王\end{CJK}/Pet Tycoon & City & 4.06M\\
        \begin{CJK}{UTF8}{gbsn}大国重工\end{CJK}/Heavy Industries of a Great Nation & City & 2.84M\\
        \begin{CJK}{UTF8}{gbsn}材料帝国\end{CJK}/Empire of Materials & City & 2.46M\\
        \begin{CJK}{UTF8}{gbsn}修真聊天群\end{CJK}/Cultivation Chat Group & City & 8.47M\\
        \begin{CJK}{UTF8}{gbsn}天启之门\end{CJK}/Door of Revelation & City & 3.05M\\
        \begin{CJK}{UTF8}{gbsn}相声大师\end{CJK}/CrossTalk Master & City & 2.30M\\
        \begin{CJK}{UTF8}{gbsn}手术直播间\end{CJK}/Surgery Livestream & City & 6.29M\\
        \begin{CJK}{UTF8}{gbsn}大医凌然\end{CJK}/Great Doctor Lingran & City & 2.99M\\
        \begin{CJK}{UTF8}{gbsn}第一序列\end{CJK}/The First Sequence & City & 2.89M\\
        \begin{CJK}{UTF8}{gbsn}鬼吹灯\end{CJK}/Ghost Blows Out the Light & City & 0.92M\\
        \begin{CJK}{UTF8}{gbsn}盗墓笔记\end{CJK}/Tomb Raider's Notes & City & 1.43M\\
        \begin{CJK}{UTF8}{gbsn}黎明医生\end{CJK}/Dawn Doctor & City & 1.82M\\
        \begin{CJK}{UTF8}{gbsn}深夜书屋\end{CJK}/Midnight Bookstore & City & 3.72M\\
        \begin{CJK}{UTF8}{gbsn}扶摇皇后\end{CJK}/Queen Fuyao & Romance & 1.34M\\
        \begin{CJK}{UTF8}{gbsn}似锦\end{CJK}/Splendid & Romance & 1.71M\\
        \begin{CJK}{UTF8}{gbsn}炮灰攻略\end{CJK}/Cannon Fodder Strategy & Romance & 4.44M\\
        \begin{CJK}{UTF8}{gbsn}花颜策\end{CJK}/Floral Beauty Tactics & Romance & 2.30M\\
        \begin{CJK}{UTF8}{gbsn}君九龄\end{CJK}/Lady Jiuling & Romance & 1.95M\\
        \begin{CJK}{UTF8}{gbsn}诛砂\end{CJK}/Sand Revenge & Romance & 1.63M\\
        \begin{CJK}{UTF8}{gbsn}凤门嫡女\end{CJK}/Phoenix Gate's Legitimate Daughter & Romance & 3.32M\\
        \begin{CJK}{UTF8}{gbsn}花繁春正茗\end{CJK}/Flourishing Sping Tea & Romance & 2.61M\\
        \begin{CJK}{UTF8}{gbsn}覆手繁华\end{CJK}/Hand Overturning Prosperity & Romance & 1.73M\\
        \begin{CJK}{UTF8}{gbsn}金陵春\end{CJK}/Jinling Spring & Romance & 1.87M\\
        \begin{CJK}{UTF8}{gbsn}慕南枝\end{CJK}/Yearning for South Branch & Romance & 2.53M\\
        \begin{CJK}{UTF8}{gbsn}一品仵作\end{CJK}/Top-Rank Coroner & Romance & 2.25M\\
        \begin{CJK}{UTF8}{gbsn}琅琊榜\end{CJK}/Nirvana in Fire & Romance & 0.74M\\
        \begin{CJK}{UTF8}{gbsn}你和我的倾城时光\end{CJK}/You Are My Glory & Romance & 0.30M\\
        \begin{CJK}{UTF8}{gbsn}挚野\end{CJK}/Wild Devotion & Romance & 0.48M\\
        \begin{CJK}{UTF8}{gbsn}中国铁路人\end{CJK}/China's Railway Man & Romance & 0.51M\\
        \begin{CJK}{UTF8}{gbsn}听说你喜欢我\end{CJK}/Heard You Love Me & Romance & 0.88M\\
        \begin{CJK}{UTF8}{gbsn}写给鼹鼠先生的情书\end{CJK}/Love Letter to Mr. Mole & Romance & 0.31M\\
        \begin{CJK}{UTF8}{gbsn}好想喜欢你\end{CJK}/Really Want to Like You & Romance & 0.17M\\
        \begin{CJK}{UTF8}{gbsn}乔先生的黑月光\end{CJK}/Mr. Qiao's Black Moonlight & Romance & 1.17M\\
        \begin{CJK}{UTF8}{gbsn}非常暖婚，我的超级英雄\end{CJK}/Very Warm Marriage: My Superhero & Romance & 0.55M\\
        \begin{CJK}{UTF8}{gbsn}金牌女律师\end{CJK}/Gold Medal Female Lawyer & Romance & 0.60M\\
        \begin{CJK}{UTF8}{gbsn}明月度关山\end{CJK}/Bright Moon Passing Mountain Pass & Romance & 0.59M\\
        \begin{CJK}{UTF8}{gbsn}玉堂留故\end{CJK}/Jade Hall Reminiscence & Romance & 0.20M\\
        \begin{CJK}{UTF8}{gbsn}凶案调查\end{CJK}/Homicide Investigation & Romance & 1.54M\\
        \begin{CJK}{UTF8}{gbsn}你好消防员\end{CJK}/Hello Firefighter & Romance & 0.52M\\
        \begin{CJK}{UTF8}{gbsn}如果深海忘记了\end{CJK}/If the Deep Sea Forgets & Romance & 0.28M\\
        \begin{CJK}{UTF8}{gbsn}谁在时光里倾听你\end{CJK}/Who Listens to You in Time & Romance & 0.36M\\
        \begin{CJK}{UTF8}{gbsn}仙灵图谱\end{CJK}/Immortal Spirit Atlas & Romance & 3.77M\\
        \begin{CJK}{UTF8}{gbsn}师父又掉线了\end{CJK}/Master Is Offline Again & Romance & 1.54M\\
        \begin{CJK}{UTF8}{gbsn}女机长\end{CJK}/Female Captain & Romance & 0.46M\\

        \bottomrule
	\end{tabularx}
	\caption{100 online novels collected by the National Library of China (part2)}
    \label{100part2}
\end{table*}%

\end{document}